\documentclass[10pt,twocolumn,letterpaper]{article}

\usepackage{cvpr}      
\usepackage{cuted}
\usepackage{float}
\usepackage{algorithm}
\usepackage{pifont} 
\usepackage{natbib}
\usepackage{algpseudocode}
\usepackage{subcaption}
\usepackage[table]{xcolor}
\usepackage{lipsum}
\usepackage{multirow}
\usepackage{placeins}
\definecolor{cvprblue}{rgb}{0.21,0.49,0.74}
\definecolor{firstred}{RGB}{255,170,170}   
\definecolor{secondyellow}{RGB}{245, 191, 66} 
\definecolor{thirdyellow}{RGB}{252, 239, 169}  

\newcommand{\hlfirst}[1]{\colorbox{cyan!28}{#1}}  
\newcommand{\hlsecond}[1]{\colorbox{cyan!10}{#1}}

\newcommand{\hllfirst}[1]{\colorbox{Goldenrod!44}{#1}}  
\newcommand{\hllsecond}[1]{\colorbox{Goldenrod!20}{#1}}

\usepackage[pagebackref,breaklinks,colorlinks,allcolors=cvprblue]{hyperref}

\newcommand{\ste}{\textcolor{black}{\mbox{\texttt{TE}}}\xspace}
\newcommand{\sre}{\textcolor{black}{\mbox{\texttt{RE}}}\xspace}

\newcommand{\vv}[1]{\mathbf{#1}}

\def\paperID{1116} 
\def\confName{CVPR}
\def\confYear{2026}
\usepackage{comment}
\usepackage{hhline}
\usepackage{boldline}
\usepackage{float}
\usepackage{pifont}

\title{DualCamCtrl: Dual-Branch Diffusion Model for \\ Geometry-Aware Camera-Controlled Video Generation }

\author{
Hongfei Zhang\textsuperscript{1}\thanks{These authors contributed equally.} \quad
Kanghao Chen\textsuperscript{1,5}\footnotemark[1] \quad
Zixin Zhang\textsuperscript{1,5} \quad
Harold H. Chen\textsuperscript{1,5} \\
Yuanhuiyi Lyu\textsuperscript{1} \quad
Yuqi Zhang\textsuperscript{3} \quad
Shuai Yang\textsuperscript{1} \quad
Kun Zhou\textsuperscript{4}\quad
Yingcong Chen\textsuperscript{1,2}\thanks{Corresponding author, yingcongchen@ust.hk} \\
[0.25cm]
\textsuperscript{1}HKUST (GZ) \quad
\textsuperscript{2}HKUST \quad
\textsuperscript{3}Fudan University \quad
\textsuperscript{4}Shenzhen University \quad
\textsuperscript{5} Knowin \\
[0.15cm]
}

\begin{document}

\maketitle

\begin{strip}
    \centering
    \centering
    \vspace{-7em}
    \includegraphics[width=\textwidth]
    {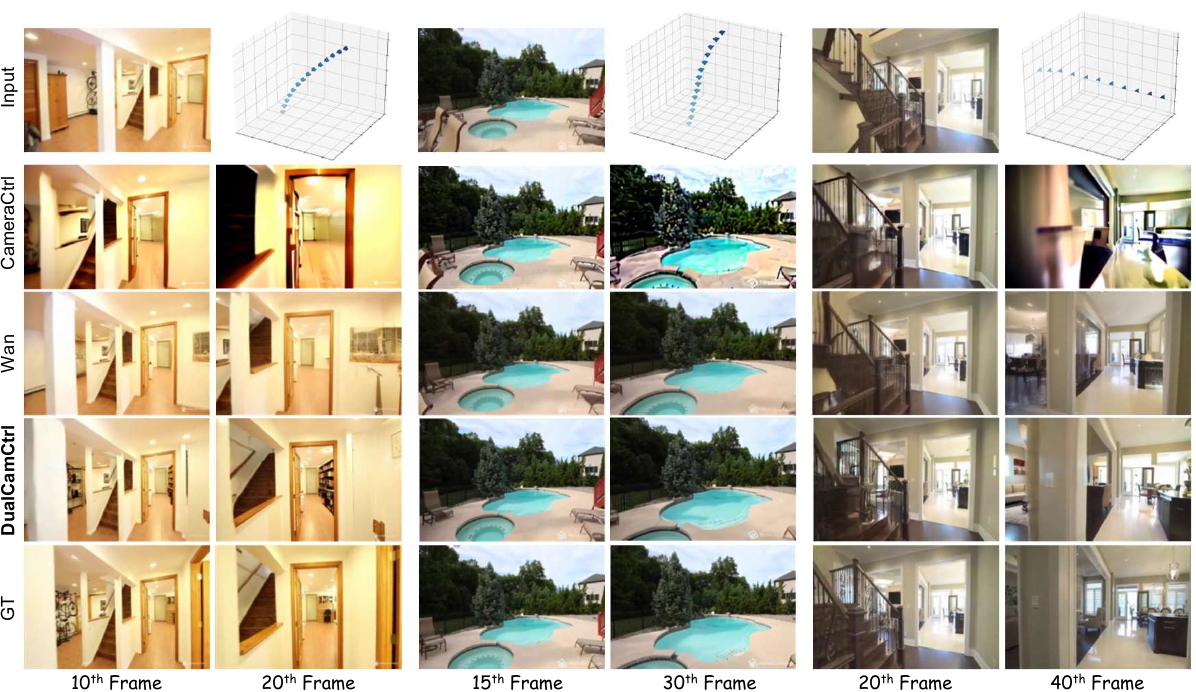}
    \vspace{-1.2em}
    \captionof{figure}{\textbf{Comparison with state-of-the-art methods~\cite{he2024cameractrl,wan2025123} on camera-controlled video generation.}
    Under an identical target camera trajectory and a single input image, our approach achieves the closest adherence to the camera motion and yields the best perceptual quality.}
    \label{fig:teaser}
    \vspace{-1.2em}
\end{strip}

\begin{abstract}
This paper presents \textbf{DualCamCtrl}, a novel end-to-end diffusion model for camera-controlled video generation. Recent works have advanced this field by representing camera poses as ray-based conditions, yet they often lack sufficient scene understanding and geometric awareness.
DualCamCtrl specifically targets this limitation by introducing a dual-branch framework that mutually generates camera-consistent RGB and depth sequences.
To harmonize these two modalities, we further propose the \textbf{S}emant\textbf{I}c \textbf{G}uided \textbf{M}utual \textbf{A}lignment (SIGMA) mechanism, which performs RGB--depth fusion in a semantics-guided and mutually reinforced manner.
These designs collectively enable DualCamCtrl to better disentangle appearance and geometry modeling, generating videos that more faithfully adhere to the specified camera trajectories.
Additionally, we analyze and  reveal the distinct influence of depth and camera poses across denoising stages and further demonstrate that early and late stages play complementary roles in forming global structure and refining local details. Extensive experiments demonstrate that DualCamCtrl achieves more consistent camera-controlled video generation \textbf{with over 40\%  reduction} on camera motion errors compared with prior methods.
\textit{Our project page: \href{https://soyouthinkyoucantell.github.io/dualcamctrl-page/}{\textcolor{purple}{https://soyouthinkyoucantell.github.io/dualcamctrl-page/}}}
\end{abstract}   
\section{Introduction}


Advances in video diffusion models~\cite{gu2025diffusion,wang2025cinemaster,guo2024i2v,ho2022video,ho2022imagen,yang2024cogvideox,liang2024movideo,wang2025lavie,guo2024sparsectrl,geng2025motion,jain2024peekaboo,koroglu2025onlyflow,bahmani2025ac3d,cai2024generative,yeh2024texturedreamer,wang2025transpixeler} have significantly improved the quality of video generation from textual descriptions. These developments have extended beyond basic text-to-video (T2V) or image-to-video (I2V) generation, with growing efforts toward integrating control mechanisms into the video generation process, such as camera control~\cite{kuang2024collaborative,bahmani2025ac3d,wang2024motionctrl,he2024cameractrl,he2025cameractrl,zhou2025stable,jin2025flovd,zheng2024cami2v,hou2024training,zhao2024motiondirector}, motion transfer~\cite{chen2023control,huang2025videomage,meral2024motionflow,zhao2024motiondirector,pondaven2025video,motamed2024investigating,wang2025tiv}, object manipulation~\cite{kahatapitiya2024object,peruzzo2024vase,saini2024invi,feng2024ccedit,tu2024motioneditor,wu2024fairy}, and multi-modality prediction~\cite{xi2025omnivdiff,pallotta2025syncvp,yang2025unified,shao2025learning,yang2024depthanyvideo,wang2025vidseg,dong2025depthsync,hu2025depthcrafter,shao2024learningtemporallyconsistentvideo}. Among them, camera control has emerged as a particularly promising direction, enabling natural camera movements and viewpoint transitions that bridge the gap between generative modeling and real-world cinematography, supporting applications from virtual directors to interactive 3D scene generation~\cite{zhou2025stable,huang2025voyager,wang2025cinemaster,wu2025difix3d+,gu2025das,chen2024cinepregen,cai2024generative}.

Building on this progress, camera-conditioned video diffusion models~\cite{wang2024motionctrl,zhou2025stable,he2024cameractrl,he2025cameractrl,bahmani2025ac3d} have been explored to enable models to respond to camera trajectories during video synthesis. However, most approaches rely solely on camera poses or further, their ray-conditioned representation (a.k.a.~pl\"ucker embedding) as control signals. 
{Although ray-conditioned models~\cite{wang2024motionctrl,bahmani2025ac3d,he2024cameractrl,zhou2025stable,huang2025voyager} encode camera motion by projecting poses into per-pixel ray directions, they inherently lack explicit scene understanding because the model must implicitly infer 3D structure to produce videos coherent with the given camera motion.}
This limitation results in suboptimal camera motion consistency, as these models are constrained by the entanglement of appearance and structure modeling and fail to fully capture the underlying geometry of the scene. 
Given these challenges, it is natural to incorporate geometric cues that encode scene structure. With advances in modern depth estimation~\cite{hu2025depthcrafter,video_depth_anything,yang2024depthanyvideo}, we turn to depth as an additional geometric modality that complements camera-pose conditioning. 

However, to the best of our knowledge, few prior works have explored incorporating depth within the context of end-to-end camera-conditioned video generation models to facilitate precise camera control. This raises the fundamental question of this paper: \textit{What is the \textbf{inherent relationship} between depth and camera control, and how can depth be effectively incorporated to further \textbf{benefit} camera-conditioned video generation?} To fill this gap, we explore an effective way to incorporate depth as an additional geometric cue into camera-conditioned video diffusion models. Yet, achieving coherent interaction between the depth and RGB modalities remains non-trivial. Given a single frame and its corresponding depth map provided by existing depth estimators~\cite{video_depth_anything}, our empirical results indicate that naive injection of depth information into existing camera-conditioned models does not consistently improve generation quality. For example, model conditioned on single-frame depth lacks sufficient temporal context to provide stable geometric cues, often leading to  unnatural scene structures (\cref{fig:single_frame_depth}).
Conversely, jointly modeling RGB and depth within a single branch introduces noticeable interference between the modalities, resulting in degraded synthesis quality (\cref{fig:single_branch_misalign}).
These observations underscore the need for an effective integration strategy that leverages depth to improve the consistency of camera motion.

To this end, we propose a dual-branch video diffusion model that effectively integrates depth information for camera control. Conditioned on shared camera poses, the model generates camera-consistent RGB and depth sequences, where depth provides guidance throughout the video, and the two modalities interact in a way that minimizes mutual interference while maintaining complementary contributions.
We further introduce the \textbf{\underline{S}}emat\textbf{\underline{I}}c \textbf{\underline{G}}uided \textbf{\underline{M}}utual \textbf{\underline{A}}lignment~(SIGMA) mechanism to harmonize semantic and geometric representation. Motivated by the principles of prioritizing semantic fidelity and enabling mutual feedback, SIGMA ensures that the RGB and depth branches evolve together, leading to more consistent and stable video generation. Additionally, a two-stage training strategy is adopted to stabilize the training process, which consists of a decoupled stage followed by a fusion stage.
Moreover, we conduct analysis and reveal that depth and camera poses affect latent representations in distinct ways across stages and layers. We also find that early and late denoising stages play complementary roles in the generation process where the former establishes global structure and the latter refines local details, offering insight for the community.



To summarize, our main contributions are threefold:

\begin{itemize}
    \item We propose \textbf{\textit{DualCamCtrl}}, a dual-branch video diffusion framework that effectively incorporates geometric cues via the  SIGMA mechanism and two-stage training.

    \item We investigate how depth and camera pose influence the RGB denoising process and the distinct roles of early and late denoising stages. This analysis sheds light on their influences and provides insights for future research.

    \item Extensive experiments validate our state-of-the-art capabilities, achieving \textbf{more than 40\% reductions} in rotation errors compared with previous methods.

\end{itemize}

\section{Related Work}
\label{sec:related}                             

\paragraph{Foundation model of video generation.}
Recent foundational diffusion-based video generation models, such as {CogVideoX}~\cite{CogVideoX}, {Stable Video Diffusion (SVD)}~\cite{SVD}, {Wan}~\cite{wan2025123}, {Lumiere}~\cite{Lumiere}, {VideoCrafter}~\cite{VideoCrafter2,chen2023videocrafter1}, {ModelScope-T2V}~\cite{ModelScopeT2V}, {Latte}~\cite{Latte}, and {Open-Sora}~\cite{OpenSoraPlan,OpenSora2}, have established powerful baselines for text-to-video and conditional video synthesis. These models employ large-scale latent diffusion or diffusion transformer architectures to achieve high temporal consistency, realistic motion, and fine-grained scene dynamics.
Building upon these foundational models, subsequent studies have explored their adaptation to various downstream video generation tasks, including camera control,~\cite{kuang2024collaborative,bahmani2025ac3d,wang2024motionctrl,he2024cameractrl,he2025cameractrl,zhou2025stable,jin2025flovd,zheng2024cami2v,hou2024training,zhao2024motiondirector}, motion transfer~\cite{chen2023control,huang2025videomage,meral2024motionflow,zhao2024motiondirector,pondaven2025video,motamed2024investigating,wang2025tiv}, object manipulation~\cite{kahatapitiya2024object,peruzzo2024vase,saini2024invi,feng2024ccedit,tu2024motioneditor,wu2024fairy}, and multi-modality prediction~\cite{xi2025omnivdiff,pallotta2025syncvp,yang2025unified,shao2025learning,yang2024depthanyvideo,wang2025vidseg,dong2025depthsync,hu2025depthcrafter,shao2024learningtemporallyconsistentvideo}.
In this work, we focus on the problem of camera-controllable video generation, aiming to enable more consistent camera control video generation built upon diffusion-based foundation models~\cite{wan2025123}.

\noindent\textbf{Camera control for video models.} 
Early efforts toward camera-controllable video generation were pioneered by methods such as MotionCtrl~\cite{wang2024motionctrl}, which explicitly encode camera poses to guide video diffusion models. Following this line, CameraCtrl~\cite{he2024cameractrl} and VD3D~\cite{bahmani2024vd3d} further introduce ray-based conditioning (\ie, Pl\"ucker embeddings~\cite{sitzmann2021light}) to inject camera parameters into pretrained diffusion backbones. More recently,~\cite{he2025cameractrl} and~\cite{wan2025123} extend this idea by incorporating camera pose information through pre-DiT conditioning. AC3D~\cite{bahmani2025ac3d} provides a deeper analysis of Video DiT behavior and mitigates training data limitations by constructing dynamic-scene datasets with rigid camera motion. Despite these advances, existing approaches still struggle to preserve the intended camera trajectory.


\noindent\textbf{Novel view synthesis (NVS) with video diffusion.}
Generating novel views from a set of posed images has seen significant progress~\cite{kerbl20233d,mildenhall2021nerf,tancik2023nerfstudio}.
With the recent success of image and video generation models, ReconFusion~\cite{wu2024reconfusion} and CAT3D~\cite{gao2024cat3d} began leveraging the prior knowledge learned by these models to facilitate sparse-view NVS. However, because they require per-scene optimization, these methods remain inherently slow. Recent works such as ReconX~\cite{liu2024reconx}, ViewCrafter~\cite{yu2024viewcrafter}, and GEN3C~\cite{ren2025gen3c} incorporate depth and point clouds to facilitate NVS and enable camera control from a few image inputs, which is somewhat similar to our setting. Nonetheless, they rely on multi-stage procedures that project and unproject point clouds into explicit camera views, in contrast to our end-to-end camera-conditional video generation. 
We pioneer the incorporation of depth into end-to-end camera-controlled video generation models to facilitate precise camera control.

\begin{figure*}[t]
    \centering
    \includegraphics[width=\textwidth]{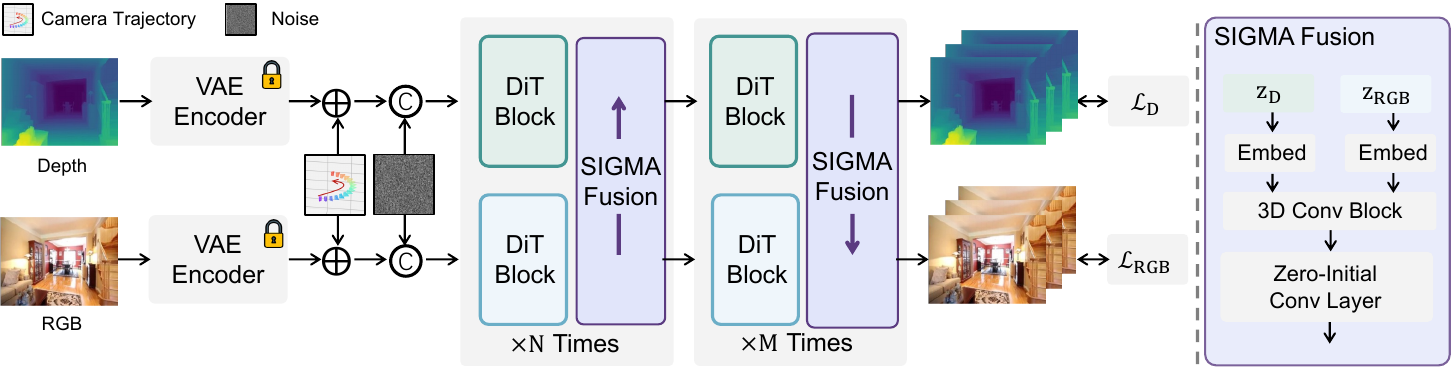}
    \vspace{-0.5em}
    \caption{
        \textbf{Overall architecture of DualCamCtrl.}
        DualCamCtrl adopts a dual-branch framework that simultaneously generates RGB and depth video latents  from an input image and its corresponding depth map. 
        The two latents are then element-wise added to the encoded Plücker embedding and concatenated with noise~(\cref{sec:dual_bracnh}). 
        Subsequently, the two modalities interact through our proposed SIGMA mechanism and fusion block (\cref{sec:pgra}). 
        During training, both predictions are supervised by their respective loss functions (\cref{sec:two_stage_training}).
    }
    \label{fig:openreels_structure}
\end{figure*}


\section{Method}
\label{sec:method}

% ...

In this section, we introduce the \textit{DualCamCtrl} framework, which enables more geometry-aware camera-conditioned video diffusion by integrating geometric cues~(\ie depth) in a dual-branch manner. 
We first provide necessary preliminaries in \cref{sec:preliminary}. In \cref{sec:dual_bracnh}, we describe how depth is injected via our {dual-branch architecture}. We introduce the \textbf{{S}}emat\textbf{{I}}c \textbf{{G}}uided \textbf{{M}}utual \textbf{{A}}lignment and 3D fusion strategy for interaction between RGB and depth branches in~\cref{{sec:pgra}}. \cref{sec:two_stage_training} details our {two-stage training pipeline}.

\subsection{Preliminary}
\label{sec:preliminary}


The video diffusion process typically involves a gradual denoising procedure, where a noisy latent representation of the video $z_t$ is progressively transformed into a high-quality video sequence. In camera-controlled video generation, the process is mainly conditioned on two inputs: the context condition \(c\) (\ie~images or textual descriptions) and the camera pose \((\mathbf{R}, \mathbf{t})\). Following recent works~\cite{wan2025123,he2024cameractrl,bahmani2025ac3d}, we encode camera geometry through ray-based conditioning, where each pixel is associated with a Plücker ray representation derived from $(\mathbf{R}, \mathbf{t})$~(See~\cref{alg:ray_condition}).

Formally, given a video $\vv{V} \in \mathbb{R}^{T \times 3 \times H \times W}$, where $T$ is the number of frames, and $H, W$ are the height and width of each frame. This video is first encoded into a latent space using a pretrained encoder $\varepsilon$, producing
$z_0 = \varepsilon(\vv{V})$, where $z_0 \in \mathbb{R}^{T^{\prime} \times C^{\prime} \times h\times w}$. 
Under our setting~\cite{wan2025123}, the latent $z$ has temporal length $T' = \frac{T-1}{4}+1$, channel dimension $C^{\prime} = 16$ and spatial dimensions $h = H/8$, $w = W/8$. The diffusion process then gradually denoises $z_t$ at each timestep $t$, conditioned on inputs $c$, which can be an image $c_{\text{image}}$ or text $c_{\text{text}}$, as well as camera pose parameters $(\mathbf{R}, \mathbf{t})$. The training objective is to predict the noise $\epsilon$ added to the latent~(\cref{eq:loss_pre}):  
\begin{equation}
\mathcal{L} = \mathbb{E}_{\epsilon \sim \mathcal{N}(0, I)}\Big[ \big\| \epsilon - \theta\big( z_t, t, c, \mathbf{R}, \mathbf{t} \big) \big\|^2 \Big],
\label{eq:loss_pre}
\end{equation}
where $\theta$ denotes the denoising network. During the inference stage, the denoising process is performed iteratively over a sequence of timesteps $t$. After denoising in the latent space, the output $\hat{z}$ is decoded by the decoder $D$ to obtain the predicted video $\hat{\vv{V}}$:
$\hat{\vv{V}} = D(\hat{z})$.

\subsection{Dual-Branch Camera-Controlled Video Generation}
\label{sec:dual_bracnh}

Depth has long been recognized as a crucial cue for scene understanding as it explicitly encodes scene geometry and spatial layout~\cite{shao2025learning,hu2025depthcrafter,lapid2023gd,choe2021volumefusion, Weder2020RoutedFusion, Tateno2017CNNSLAM,Menini2021Joint3DReconstruction, Laidlow2022DeepFusion,Rosinol2022ProbVolFusion, Gupta2020ImprovedMultiviewDepth}. However, despite the importance of geometric priors, the integration of depth cues in camera-conditioned video generation remains unexplored. In this section, we investigate how depth information can be leveraged to enhance camera-conditioned video synthesis. Specifically, under the I2V setting, the input consists of a single reference frame, and monocular depth estimator~\cite{video_depth_anything} allows us to predict a plausible depth map for that frame. However, relying solely on this single-frame depth is insufficient for guiding long-range video generation, as it provides only a static snapshot of the scene's geometry without maintaining temporal consistency. 
Moreover, naively coupling the generation of RGB and depth modalities tends to cause interference between them~(\cref{fig:single_frame_depth,fig:single_branch_misalign}).
This motivates us to design a framework that can propagate and effectively utilize depth information throughout the video sequence, thereby assisting  RGB synthesis while minimizing interference. To this end, we make a trial to adopt a dual-branch architecture, where RGB and depth features are processed in parallel, each conditioned on the same camera poses to generate modality-specific representations (\cref{fig:openreels_structure}). This design allows the model to produce coherent information for both modalities under consistent camera motion. 

 
Formally, given an input image $\mathbf{I_{\text{RGB}}} \in \mathbb{R}^{3 \times H \times W}$, we first employ a monocular depth estimator~\cite{video_depth_anything} to obtain its depth map $\mathbf{I_{\text{D}}} \in \mathbb{R}^{1 \times H \times W}$. Following previous works~\cite{he2024lotus,hu2025depthcrafter,shao2024learningtemporallyconsistentvideo,yang2024depthanyvideo}, the estimated depth is replicated along the channel dimension to match the channel number of the RGB input. Both the RGB image and the replicated depth are then encoded into the latent space using a pretrained encoder $\varepsilon$, producing their respective latent representations $z^{\mathbf{I}_\text{RGB}}$ and $z^{\mathbf{I}_\text{D}}$.
The two latent representations are both zero-padded in the frame channel to match the length of noise. Then they are element-wisely added with the same encoded pl\"ucker embedding which represents the camera pose $(\mathbf{R}, \mathbf{t})$. Afterwards, they are concatenated with the same Gaussian noise $\epsilon$ and processed by a denoising network $\theta$~(\ie Diffusion Transformer~\cite{peebles2023scalable}), which consists of two parallel branches corresponding to the RGB and depth modalities. The model $\theta$ jointly predicts the denoised latent sequences, yielding an RGB video latent $\hat{z}^{\vv{V}_\text{RGB}}$ and a depth video latent $\hat{z}^{\vv{V}_\text{D}}$.

During training, the predicted latents are supervised by the ground-truth video latents ${z}^{\vv{V}_\text{RGB}}$ and ${z}^{\vv{V}_\text{D}}$, which are obtained by encoding the ground-truth RGB-D video $\vv{V}^{\text{RGB}}$ and $\vv{V}^{\text{D}}$. During inference, a pretrained decoder $\boldsymbol{D}$ is employed to reconstruct the corresponding RGB from the predicted latents. The training strategy is further described in~\cref{sec:two_stage_training}.
Details of the T2V setting are provided in the supplementary material due to space limitations~(\cref{fig:t2v_arch}).

        

\subsection{Injecting Depth Cues for Mutual Alignment} 
\label{sec:pgra}

\begin{figure*}[t]
    \centering
    \includegraphics[width=\linewidth]{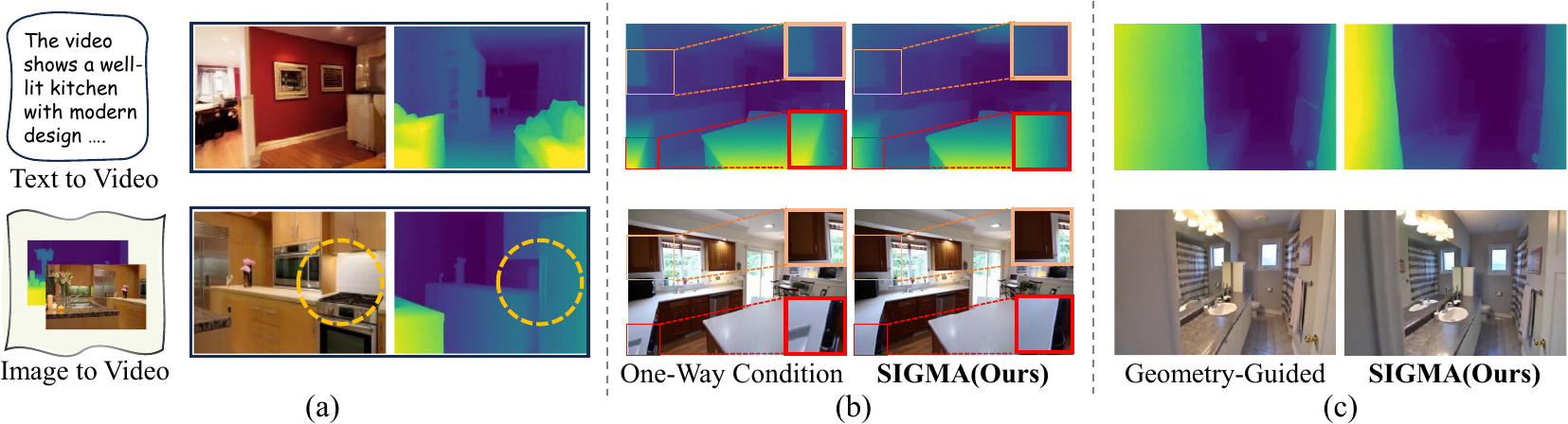}
    \vspace{-1.5em}
    \caption{\textbf{(a) Illustration of modality misalignment.} Independent RGB and depth latent evolution leads misalignment across frames. This motivates the design of our SIGMA strategy to establish coherent cross-modal alignment. \textbf{(b) Comparison with one-way alignment.}  One-way alignment transfers information unidirectionally, leading to misalignment on local semantics.
        {\textbf{(c) Comparison with geometry-guided alignment}} Under the geometry-guided setting, geometry cues evolved too quickly and become inconsistent with RGB motion.}
    \vspace{-1em}
    \label{fig:misalignment}
\end{figure*}

\paragraph{{{S}}emat{{i}}c {{G}}uided {{M}}utual {{A}}lignment.}
Based on our dual-branch framework, one branch synthesizes the RGB sequence, while the other generates the corresponding depth sequence. Despite being conditioned on the same input~(\ie image or text), the two branches evolve independently, resulting in  inconsistency between their outputs, which necessitates an effective alignment strategy~(\cref{fig:misalignment}).

However, simple fusion strategies remain insufficient.
We observe that \textit{one-way alignment} transferring features from one modality to another without feedback, fails to preserve semantic consistency, while \textit{geometry-guided alignment} overemphasizes geometric cues and disrupts appearance coherence. To overcome this, we introduce \textbf{{S}}emat\textbf{{I}}c \textbf{{G}}uided \textbf{{M}}utual \textbf{{A}}lignment~(SIGMA). SIGMA adopts a semantic-guided bidirectional design: early layers leverage RGB features to anchor semantic structure, while later layers incorporate depth feedback to refine geometry.

Notably, this design is motivated by two key insights. \textbf{\ding{182} Priority matters:} Appearance should first dominate the initial stages to ensure semantic fidelity, while depth is introduced later as a complementary corrective signal that refines geometric consistency. \textbf{\ding{183} Mutual feedback matters:} Enabling both branches to inform each other avoids the imbalance of one-way alignment, leading to more consistent representation. Together, these principles enable SIGMA to more effectively harmonize semantic and geometric representations, ultimately yielding more visually coherent video generation~(\cref{tab:ablation_pgra}).


\begin{figure}[t]
    \centering
    \includegraphics[width=\linewidth]{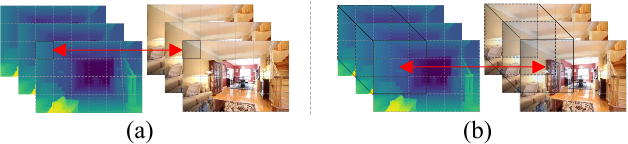}
    \includegraphics[width=\linewidth]{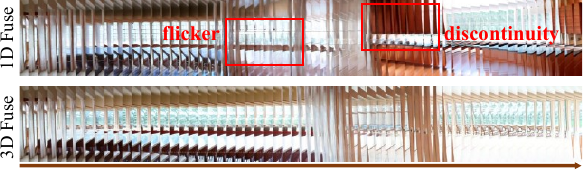}
    \vspace{-1em}
    \caption{
        \textbf{Comparison of fusion strategies.} 
                (1.a) Previous shallow linear fusion operates in a pixel-wise manner, ignoring temporal and spatial context and causing inconsistency over time.
        (1.b) We introduce {3D Fusion strategy} to extend the fusion formulation from 1D to 3D by incorporating 3D operations.
        (2) Traditional linear-layer fusion often leads to visual artifacts, while our methods produce smoother, more coherent results.    }
    \vspace{-1em}
    \label{fig:3dfusion}
\end{figure}



\vspace{-1.2em}
\paragraph{3D Fusion Strategy.}
Our experiments reveal that achieving temporal and spatial consistency between RGB and depth sequences cannot rely solely on alignment strategies~(\cref{fig:3dfusion}). While SIGMA effectively coordinates the two branches, conventional shallow projection layers~(\textit{e.g.}, linear feature injection as in~\cite{gu2025das,zheng2024cami2v,wang2025cinemaster}) still ignore the temporal ordering and spatial layout of video, treating all latent pixels uniformly across frames and often producing temporally inconsistent outputs.
To effectively capture spatiotemporal cues, we introduce 3D convolutions into the fusion block, enabling the model to fuse semantic and geometry information over both spatial and temporal dimension~(\cref{fig:openreels_structure}). To balance computational cost and parameter efficiency, we further adopt an efficient bottleneck design inspired by~\cite{sandler2018mobilenetv2,howard2017mobilenets,tran2018closer}. A frame-wise gating mechanism is involved to adaptively modulate the fusion strength according to temporal dynamics. We provide a more detailed description in the appendix~(\cref{fig:3dfusion_detail}).

\subsection{Two-stage Training Pipeline}
\label{sec:two_stage_training}
 Recall that our training objective is twofold: enable each modality to develop generative competence and foster effective cross-modal interaction. To achieve this, we adopt a carefully staged schedule that balances the learning dynamics of both modalities.
 We first observe that the RGB branch, initialized from pretrained visual backbones, naturally possesses a strong prior for appearance generation, while the depth branch starts from scratch without generative capability of depth modality. Directly training both branches jointly leads to severe convergence issues, where the depth stream fails to learn meaningful geometry and, in turn, destabilizes the overall optimization~(\cref{fig:two_visual_comparison,table:ablation_twostage}). Thus, to perform effective training, we adopt a two-stage strategy: first, we independently learn appearance and geometry in the decoupling stage, and then we enable cross-branch interaction through a fusion block in the fusion stage. The detailed process is described below.

\vspace{-1.2em}
\paragraph{Initialization and Decoupled Stage.}
To incorporate knowledge from pretrained diffusion models, we first initialize both the RGB and depth branches with the same model weights~\cite{wan2025123}. The primary goal of this stage is to enable the model to learn a robust depth synthesis module that can be used as a complementary signal for the RGB branch. However, no existing dataset provides simultaneously scene-level realistic video motion, corresponding camera parameters, and ground-truth depth~\cite{wu2023omniobject3d, downs2022google, mildenhall2019local, jensen2014large, barron2021mip, ling2024dl3dv, knapitsch2017tanks}. To address this, we employ state-of-the-art monocular depth estimation methods to predict depth maps for all frames and use them as supervision for the depth branch~\cite{video_depth_anything}. To avoid interference between the two modalities, we do not perform any cross-branch fusion at this stage. This separation ensures that each branch can independently capture its respective cues, appearance for RGB and geometry for depth, without destabilizing the training process~(\cref{fig:two_visual_comparison,table:ablation_twostage}).  Interestingly, we observed that although the depth branch cannot generate the target sequence initially, it still interprets depth as a hazy image and uses this as a condition for generation~(\cref{fig:misleading_generation}).

\vspace{-1.2em}
\paragraph{Fusion Stage.}
In the second stage, we enable fusion between the RGB and depth branches to exploit their complementary strengths: the RGB branch provides rich appearance cues, while the depth branch conveys geometric structure. The training objective remains a combination of depth and RGB losses~(\cref{eq:overall_loss}). We zero-initialize the fusion block, allowing their influence to emerge gradually.

\vspace{1pt}
\noindent\textbf{Loss Function.}
Let \(\gamma \in \{0, 1\}\) indicate whether cross-branch fusion is enabled, where \(\gamma = 0\) corresponds to the decoupled stage and \(\gamma = 1\) corresponds to the fusion stage. We define the 3D-aware cross features \(h_t^{\mathrm{RGB \rightarrow D}}\) (from RGB to Depth) and \(h_t^{\mathrm{D \rightarrow RGB}}\) (from Depth to RGB), with the corresponding losses for each branch as follows~(\cref{eq:rgb_loss,eq:depth_loss}):
\begin{equation}
\label{eq:rgb_loss}
\begin{split}
\mathcal{L}_{\mathrm{RGB}} = 
\mathbb{E}\Big[ \big\| 
v_t^{\mathrm{RGB}} - 
\theta_{\mathrm{RGB}}(& z_t^{\mathrm{RGB}}, t, c, \\
& \mathbf{R}, \mathbf{t}; \gamma\, h_t^{\mathrm{D \rightarrow RGB}} )
\big\|^2 \Big],
\end{split}
\end{equation}
\begin{equation}
    \label{eq:depth_loss}
    \mathcal{L}_{\mathrm{D}} = 
    \mathbb{E}\left[
    \left\| 
    v_t^{\mathrm{D}} - 
    \theta_{\mathrm{D}}\left( z_t^{\mathrm{D}}, t, c, \mathbf{R}, \mathbf{t}; \gamma\, h_t^{\mathrm{RGB \rightarrow D}} \right)
    \right\|^2
    \right].
\end{equation}
The overall loss is then defined as~(\cref{eq:overall_loss}):
\begin{equation}
    \mathcal{L}_{\mathrm{Overall}} = \mathcal{L}_{\mathrm{RGB}} + \lambda \mathcal{L}_{\mathrm{D}}.
\label{eq:overall_loss}
\end{equation}

\section{Analysis}

\label{sec:analysis}

We first conduct an analysis to gain a deeper understanding of how camera pose and depth representation influence the denoising process. Specifically, we employ {Centered Kernel Alignment} analysis~\cite{kornblith2019similarity} to quantify the relationship between RGB latent representations and various signals (\ie depth supervision and Plücker-based ray representation) throughout the denoising process. Based on this, we further investigate the relative importance of each stage in camera-conditioned video generation by analyzing common inference-time scheduling strategies and their effects on temporal coherence and camera controllability. For simplicity, we define the early stage as $t > 0.9T$, the mid stage as $0.75T < t \leq 0.9T$, and the late stage as $t \leq 0.75T$, where $T$ denotes the total number of denoising timesteps.

\begin{figure}[t]
    \centering
    \includegraphics[width=\linewidth]{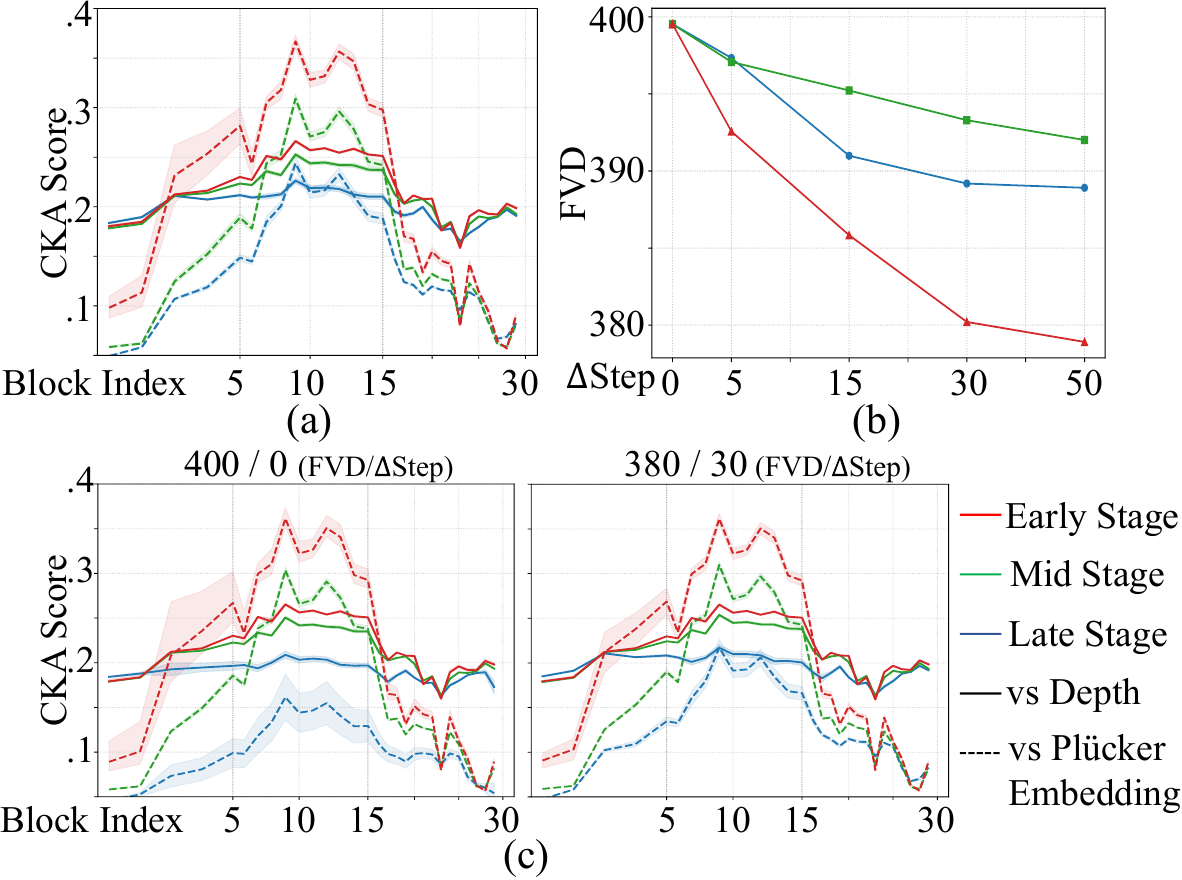}
    \vspace{-2em}
    \caption{\textbf{CKA \textit{vs.}\ pose; early-stage effects.}
        (a) RGB branch shows strongest camera motion alignment in early–mid layers.
        (b) Extra early steps yield the largest quality gains.
        (c) Stronger early-stage weights lower CKA variance and improve FVD.}
    \label{fig:analysis}
    \vspace{-1em}
\end{figure}

\subsection{Influence Dynamics Across Denoising Stages.}

CKA curves~(\cref{fig:analysis}a) indicate that the similarity between RGB latents and the Pl\"ucker ray embeddings peaks in the \emph{early-to-mid} part of the schedule (\(t \in (0.75T,T]\)), showing that camera-pose conditioning exerts its strongest influence when global structure is being formed. The similarity then drops sharply in the \emph{late} range (\(t \le 0.75T\)). Consistent with prior observations~\cite{bahmani2025ac3d,CogVideoX,wan2025123}, the \emph{mid} layers carry the most accurate information about camera pose; notably, this also holds for depth information. We further observe pronounced variability in the early-stage similarity between Pl\"ucker embeddings and RGB features, suggesting that seemingly small changes to the schedule or conditioning strength during this window can produce large effects on the final trajectory and structure. Motivated by these findings, the next subsection analyzes how the denoising stage impacts camera-controlled video generation.

\subsection{Which stage is important for camera control?}

Building on the above analysis, we ask which denoising stage contributes most to camera control video generation. We compare different timestep schedules~(\ie~more steps on \emph{early}, \emph{mid} or \emph{late}) and quantify their effects on video generation. Timesteps are sampled linearly within each predefined stage, as variations in the within-stage distribution showed negligible impact in our pilot experiments. We use 15 steps uniformly distributed across the three stages, while $\Delta$step denotes the additional steps allocated.

\vspace{-1em}
\paragraph{I. Role of Early Denoising Steps in Establishing Camera Geometry.} We observe that the {early denoising stage ($t>0.9T$)} plays a particularly critical role in defining the global camera geometry. During this period, the Plücker embeddings exhibit high variance within the RGB latent space, suggesting that the model is actively aligning its internal representation with the camera pose. Increasing the number of denoising steps in different diffusion stages improves generation quality overall (Fig.~\ref{fig:analysis}b), but the most substantial gains arise from the early stage. This suggests that additional early denoising allows the model to better consolidate the global camera structure before appearance refinement begins. In contrast, insufficient early denoising leads to unstable alignment and degraded FVD performance, indicating that the camera–scene relationship is primarily established at the start of the diffusion process~(See~\cref{fig:analysis}c).

\begin{table*}[t]
    \centering\small
    \caption{\textbf{Quantitative comparisons on I2V setting.} $\uparrow/\downarrow$ denotes higher/lower is better. We highlight the \hlfirst{best} and \hlsecond{second best}. }
    \vspace{-0.7em}
    \setlength{\tabcolsep}{3pt}
    \resizebox{1\linewidth}{!}{
    \begin{tabular}{l|cc|ccc|cc|cc|ccc|cc}
        \toprule
        \multirow{2}{*}{\textbf{Method}}& \multicolumn{7}{c|}{\textbf{RealEstate10K}} & \multicolumn{7}{c}{\textbf{DL3DV}} \\
        \cmidrule(lr){2-8} \cmidrule(lr){9-15}
         & FVD $\downarrow$ & FID $\downarrow$  & CLIPSIM $\uparrow$ & FC $\uparrow$  & MS $\uparrow$  & \sre $\downarrow$ & \ste $\downarrow$
               & FVD $\downarrow$ & FID $\downarrow$  & CLIPSIM $\uparrow$ & FC $\uparrow$  & MS $\uparrow$  & \sre $\downarrow$ & \ste $\downarrow$ \\
        \midrule
        MotionCtrl \cite{wang2024motionctrl} & 137.4 & {71.70} & 0.2401 & {0.9621} & 0.9733 & 2.80 & 1.04
                   & 158.5 & {98.51} & 0.2009 & {0.9386} & 0.9360 & 1.10 & 1.73 \\
        CameraCtrl \cite{he2024cameractrl}& 118.7 & \hlsecond{69.90} & 0.2358 & \hlsecond{0.9623} & {0.9860} & {2.38} & {1.03}
                   & 
                   {144.4} & {85.37} & 0.2015 & \hlsecond{0.9569} & \hlsecond{0.9745} & {1.04} & {1.71} \\
        Seva \cite{zhou2025stable}     & \hlsecond{104.2} & 76.69 & \hlfirst{0.3059} & 0.9570 & 0.9815 & 2.62 & \hlsecond{0.32}
                   & 206.0 & 130.9 & \hlsecond{0.2044} & 0.9443 & 0.9471 & 1.06 & {1.67} \\
        Wan \cite{wan2025123}       & {109.2} & 77.80 & {0.2859} & 0.9567 & \hlsecond{0.9863} & \hlsecond{2.08} & \hlsecond{0.32}
                   & \hlsecond{127.9} & \hlsecond{70.77} & {0.2022} & {0.9512} & {0.9731} & \hlsecond{1.01} & \hlsecond{1.64} \\
        \hline
        \textbf{DualCamCtrl}     & \hlfirst{80.38} & \hlfirst{49.85} & \hlsecond{0.2998} & \hlfirst{0.9677} & \hlfirst{0.9886} & \hlfirst{1.25} & \hlfirst{0.23}
                   & \hlfirst{92.2} & \hlfirst{53.89} & \hlfirst{0.2047} & \hlfirst{0.9637} & \hlfirst{0.9763} & \hlfirst{0.88} & \hlfirst{1.39} \\
        \bottomrule
    \end{tabular}
    }
    \vspace{-1em}
    \label{table:benchmarkcomp_i2v_combined}
\end{table*}

\vspace{-1em}
\paragraph{II. Late-stage Refinement Also Contributes to Pose-aware Detail Consistency.}
While the {early} denoising stage primarily establishes the global camera geometry, we empirically find that the {late} stage also plays a crucial role in maintaining pose-consistent local refinement. In particular, once early denoising has sufficiently stabilized the coarse spatial structure aligned with the camera trajectory~(50 steps  under our setting), the benefit of additional camera-control guidance in the early steps diminishes. By contrast, allocating more steps to the {late} stage yields larger gains where subsequent denoising helps refine object boundaries and high-frequency textures~(\cref{fig:late_artifact}). Due to space limitation, we provide more analysis in the appendix~(\cref{appendix:findings_and_analysis}).

\begin{figure}[t]
    \centering
    \includegraphics[width=\linewidth]{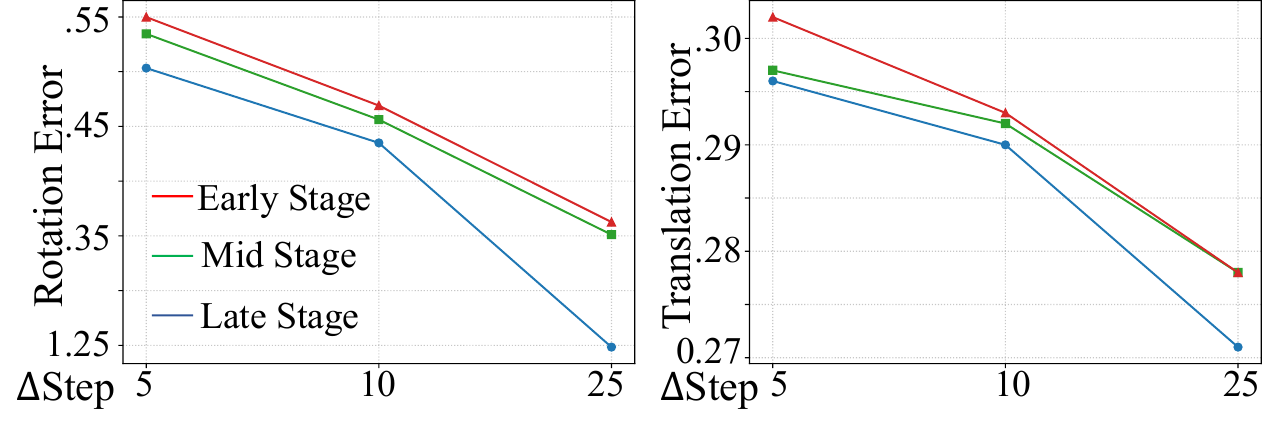}
    \includegraphics[width=\linewidth]{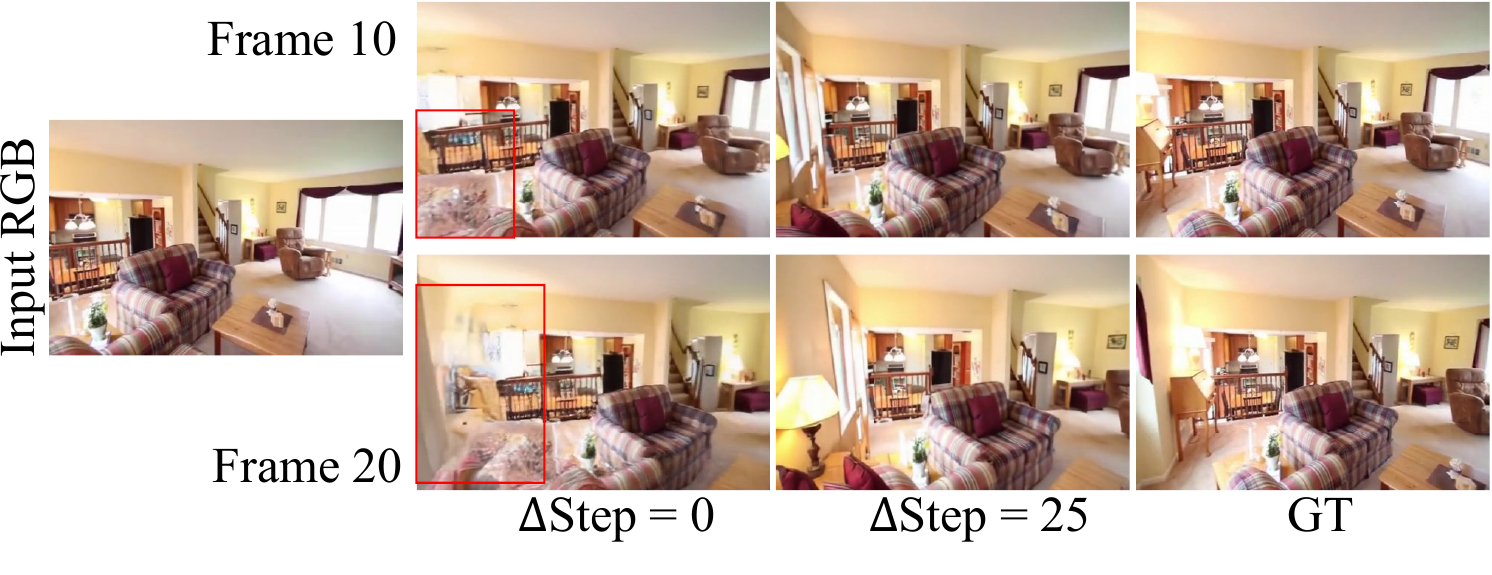}
    \vspace{-2em}
    \caption{
        \textbf{Importance of late denoising stages.}
        (1) Rotation/Translation error \textit{vs.}\ number of denoising steps under different strengths: once early steps pass a threshold, adding more early steps yields diminishing returns in geometric consistency, whereas late steps are key for finer appearance fidelity.
        (2) Late denoising sharpens object boundaries and surface details, indirectly improving pose-aware consistency. Please zoom in for better view.
    }
    \vspace{-1em}
    \label{fig:late_artifact}
\end{figure}

\begin{figure*}[t]
    \centering
    \vspace{-0.4em}
    \includegraphics[width=\textwidth]{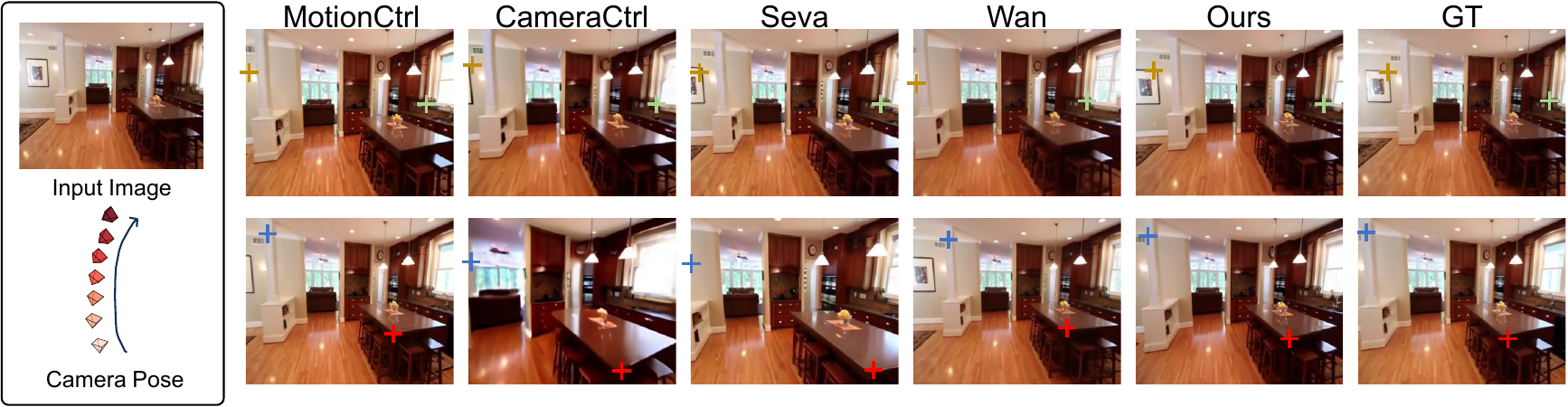}
    \vspace{-1.7em}
    \caption{
        \textbf{Comparison between our method and other state-of-the-art approaches.} 
        Given the same camera pose and input image as generation conditions, our method achieves the best alignment between camera motion and scene dynamics, producing the most visually accurate video. 
        The '$+$' signs marked in the figure serve as \textbf{anchors to indicate specific reference points for better visual comparison}. 
    }
    \vspace{-0.5em}
    \label{fig:i2v_compare}
\end{figure*}

\section{Experiments}
\label{sec:exp}


\paragraph{{Baselines.}} 
We evaluate our method against several state-of-the-art camera-controlled video generation models.
In the I2V setting, we compare our approach with \textsc{MotionCtrl}~\cite{wang2024motionctrl} and \textsc{CameraCtrl}~\cite{he2025cameractrl}, as well as two recent strong baselines, \textsc{Wan-V2.1}~\cite{wan2025123} and \textsc{SEVA}~\cite{zhou2025stable}.
In the T2V setting, we compare with \textsc{CameraCtrl}, \textsc{MotionCtrl}, and the most recent \textsc{AC3D}~\cite{bahmani2025ac3d}.
All baseline models are evaluated under the same split and evaluation metrics for a consistent and unbiased comparison. We do not compare with~\cite{luo2025camclonemaster} due to different settings and application scenarios.

\vspace{-1em}
\paragraph{Implementation Details.}
Following~\cite{bahmani2025ac3d,he2025cameractrl,wang2024motionctrl,zhou2025stable}, we train our model on the \textsc{RealEstate10K}~\cite{46965}. It contains over $60$ M video clips with per-frame camera parameters. We sample frames with a stride ranging from 0 to 8 to capture a wide range of camera motions. Our video diffusion models are developed by fine-tuning pre-trained video diffusion models~\cite{wan2025123}. We follow the original training scheme of the baseline model with 1000 denoising steps, using flow matching~\cite{lipman2022flow} as the training target.

\vspace{-1em}
\paragraph{Evaluation Metrics.}
We assess our video generation performance using several widely adopted metrics, including Fréchet Video Distance (FVD)~\cite{unterthiner2019fvd}, Fréchet Inception Distance (FID)~\cite{unterthiner2018towards}, and CLIP similarity (CLIPSim)~\cite{radford2021learning}. We also employ large-scale, open-source evaluation frameworks to evaluate Frame Consistency and Motion Strength (denoted as FC and MS)~\cite{huang2024vbench}. In addition, we calculate the rotation and translation error~(denoted as \sre and \ste) using~\cite{wang2023vggsfm} with fine-tracking. We further conduct a human study to obtain qualitative feedback on various subjective aspects~(\cref{table:user_study_results}), offering stronger evidence of the effectiveness of our method.  More training and evaluation details are provided in the appendix.

\begin{table*}[t]
    \centering\small
    \caption{\textbf{Quantitative comparisons on T2V setting across \textsc{RealEstate10K} and \textsc{DL3DV}.}}
    \vspace{-0.3em}
    \setlength{\tabcolsep}{3pt}
    \resizebox{1\linewidth}{!}{
    \begin{tabular}{l|c|ccc|cc|c|ccc|cc}
        \toprule
        \multirow{2}{*}{\textbf{Method}} 
        & \multicolumn{6}{c|}{\textbf{RealEstate10K}} 
        & \multicolumn{6}{c}{\textbf{DL3DV}} \\
        \cmidrule(lr){2-7} \cmidrule(lr){8-13}
        & FVD $\downarrow$ & CLIPSIM $\uparrow$ & FC $\uparrow$ & MS $\uparrow$ & \sre $\downarrow$ & \ste $\downarrow$ 
        & FVD $\downarrow$ & CLIPSIM $\uparrow$ & FC $\uparrow$ & MS $\uparrow$ & \sre $\downarrow$ & \ste $\downarrow$ \\
        \midrule
        MotionCtrl \cite{wang2024motionctrl} & 506.9 & {0.2744} & 0.9471 & 0.9734 & 2.90 & 1.06 
                   & 746.7 & {0.2033} & 0.9448 & 0.9462 & 1.17 & 3.23 \\
        CameraCtrl \cite{he2024cameractrl} & {426.8} & 0.2734 & \hlsecond{0.9526} & {0.9750} & {2.68} & {0.98} 
                    & {675.5} & 0.2027 & {0.9518} & {0.9610} & {1.01} & {4.00} \\
        AC3D \cite{bahmani2025ac3d} & \hlsecond{415.6} & \hlsecond{0.3044} & {0.9496} & \hlsecond{0.9811} & \hlsecond{2.67} & \hlsecond{0.38} 
             & \hlsecond{452.8} & \hlsecond{0.2078} & \hlsecond{0.9529} & \hlfirst{0.9869} & \hlsecond{1.06} & \hlsecond{2.11} \\
        \hline
        \textbf{DualCamCtrl } & \hlfirst{408.1} & \hlfirst{0.3154} & \hlfirst{0.9540} & \hlfirst{0.9918} & \hlfirst{1.23} & \hlfirst{0.25} 
                       & \hlfirst{427.4} & \hlfirst{0.2090} & \hlfirst{0.9544} & \hlsecond{0.9807} & \hlfirst{0.83} & \hlfirst{1.53} \\
        \bottomrule
    \end{tabular}}
    \vspace{-1em}
    \label{table:benchmarkcomp_t2v_combined}
\end{table*}

\subsection{Comparisons}

\paragraph{Quantitative results.}
As shown in Tables~\ref{table:benchmarkcomp_i2v_combined} and~\ref{table:benchmarkcomp_t2v_combined}, our method achieves consistent improvements across both I2V and T2V settings, demonstrating superior performance on objective metrics. Notably, due to the significant reduction in rotation and translation errors~(\ie~40\%), our method also achieves a substantial decrease in FVD, which is particularly noticeable under the I2V setting.

\vspace{-1em}
\paragraph{Qualitative results.}
As illustrated in~\cref{fig:i2v_compare}, our approach produces videos that exhibit higher visual fidelity and temporal coherence. Moreover, our method obtained the highest overall preference rate in the human study~(\cref{table:user_study_results}) confirming that the integration of depth-aware geometry yields results that are not only quantitatively superior but also subjectively more visually appealing to human observers. 
Please refer to the appendix for more results.


\subsection{Ablation Study}

In this section, we conduct ablation study to evaluate the contributions of key components in our \textit{DualCamCtrl} framework on the \textsc{Realestate10K} dataset. We put more ablation cases in the appendix~(\cref{appendix:more_ablation}).

\begin{table}[t]
    \centering\small
    \caption{\textbf{Ablation on two-stage training.}}
    \vspace{-0.7em}
    \setlength{\tabcolsep}{4pt}
    \begin{tabular}{l|cc|cc}
        \toprule
        Variant & FVD $\downarrow$ & FID $\downarrow$ & MS $\uparrow$ & FC $\uparrow$ \\
        \midrule
        Single-Stage & 96.7 & 58.3 & 0.9859 & 0.9537 \\
        \textbf{Two-Stage (Ours)} & \hllfirst{80.4} & \hllfirst{49.9} & \hllfirst{0.9886} & \hllfirst{0.9677} \\
        \bottomrule
    \end{tabular}
    \label{table:ablation_twostage}
\end{table}


\begin{figure}[t]
    \centering
    \includegraphics[width=\linewidth]{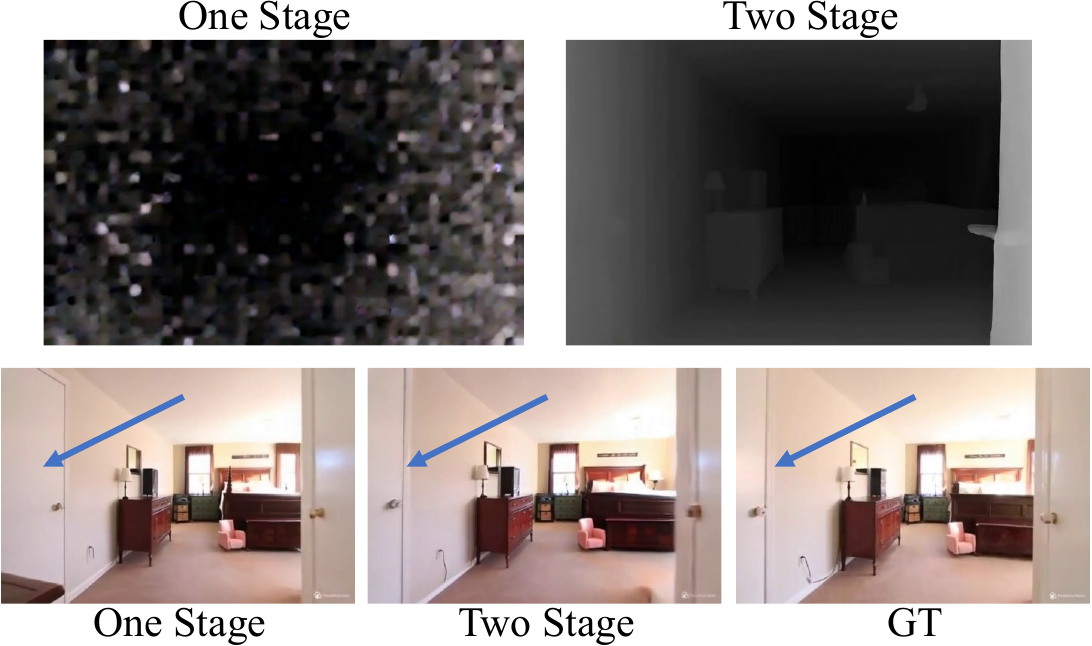}
    \vspace{-1.2em}
    \caption{
        {\textbf{Effect of Two-Stage Training:} \textit{Top}: The single-stage model fails to converge properly. \textit{Bottom}: Consequently, the geometry alignment is suboptimal, highlighting the importance of disentangling appearance and geometry learning in the two-stage training approach.}
    }
    \vspace{-1em}
    \label{fig:two_visual_comparison}
\end{figure}

\vspace{-1em}
\paragraph{Effect of Incorporating Depth.}
To assess the benefit of depth guidance, we compare our model with and without depth guidance. 
As shown in \cref{table:ablation_depth}, removing depth leads to a notable degradation in FVD and MS, indicating weaker temporal stability and geometry understanding. 
This confirms that depth provides strong structural cues for camera controlled scene generation.

\vspace{-1em}
\paragraph{Effect of Two-Stage Training.}
We further evaluate the effectiveness of our two-stage training scheme. 
As seen in \cref{table:ablation_twostage}, direct joint training (only fusion stage) leads to unstable convergence and suboptimal temporal coherence~(\cref{fig:two_visual_comparison}). 
The two-stage variant achieves superior results across all metrics, suggesting that decoupling appearance and geometry learning before fusion significantly stabilizes training and enhances cross-modal consistency.

\begin{table}[t]
    \centering\small
    \caption{\textbf{Ablation on depth conditioning.}}
    \vspace{-0.7em}
    \setlength{\tabcolsep}{4pt}
    \begin{tabular}{l|cc|cc}
        \toprule
        Variant & FVD $\downarrow$ & FID $\downarrow$ & MS $\uparrow$ & FC $\uparrow$ \\
        \midrule
        w/o Depth & 96.3 & 66.1 & 0.9872 & 0.9610 \\
        \textbf{w/ Depth (Ours)} & \hllfirst{80.4} & \hllfirst{49.9} & \hllfirst{0.9886} & \hllfirst{0.9677} \\
        \bottomrule
    \end{tabular}
    \label{table:ablation_depth}
\end{table}

\section{Conclusion}
\label{sec:conclusion}
We presented \textbf{\textit{DualCamCtrl}}, a dual-branch video diffusion model that integrates depth for more accurate camera-controlled video generation. By introducing the SemantIc Guided Mutual Alignment (SIGMA) mechanism and a two-stage training process, \emph{DualCamCtrl} effectively synchronizes RGB and depth sequences, improving geometry awareness. Our experiments show over 40\% reduction in rotation errors compared to previous methods, offering new insights into camera-controlled video generation and hopefully benefiting other video generation tasks.


\clearpage
{
    \small
    \bibliographystyle{plain}
    \bibliography{main}
}
\clearpage
\clearpage
\setcounter{page}{1}
\maketitlesupplementary

\section{Further Details of the DualCamCtrl Framework}

\subsection{Detail of 3D Fusion Block}
\begin{figure}[htbp]
    \centering
    \includegraphics[width=\linewidth]{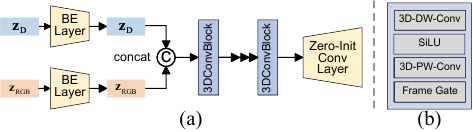}
    \vspace{-1.2em}
    \caption{\textbf{Architecture of 3D-Aware Fusion Block.} (a) Design of the fusion block. The fusion block consists of a bottleneck embedding, a series of 3D convolutional layers, a zero-initialized convolutional output layer. 
    (b) Detail of the 3D convolutional blocks. Here, {BE} stands for bottleneck embedding. {DW-Conv} stands for depthwise 3D convolution. {PW-Conv} stands for pointwise 3D convolution. Please zoom in for a better view.}
    \vspace{-1em}
    \label{fig:3dfusion_detail}
\end{figure}

To balance accuracy and efficiency, our 3D-aware fusion block adopts a bottleneck design. Embeddings are first projected into a lower-dimensional space, aggregated by stacked 3D convolutions with a depthwise, pointwise decomposition \cite{sandler2018mobilenetv2,howard2017mobilenets,tran2018closer}, and then mapped back through a zero-initialized output layer. This initialization makes the block behave like an identity at the start of training, while a frame-level gate modulates how much geometric information is injected per frame (\cref{fig:3dfusion_detail}).

Concretely, given a sequence embedding ${z}{\in}\mathbb{R}^{B\times L\times C_{\text{emb}}}$ with $L{=}T^{\prime} {\times} h {\times} w$, we reshape it to $\mathbb{R}^{B{\times} C_{\text{emb}} {\times} T^{\prime} {\times} h {\times} w}$. A bottleneck  convolution reduces the channel dimension to $C_b{=}\tfrac{C_{\text{emb}}}{4}$. The features are then processed by a stack of 3DConv blocks, each consisting of a depthwise $3{\times}3{\times}3$ convolution followed by a pointwise $1{\times}1{\times}1$ convolution. A zero-initialized $1{\times}1{\times}1$ convolution restores the channels to $C$. Finally, a frame-level gating mechanism produces $g_t\in[0,1]$ and modulates the output:
\[
\mathbf{Y}_t \;=\;  g_t \,\odot\, \mathcal{F}(\mathbf{X}_t),
\]

\begin{figure}[htbp]
    \centering
    \includegraphics[width=\linewidth]{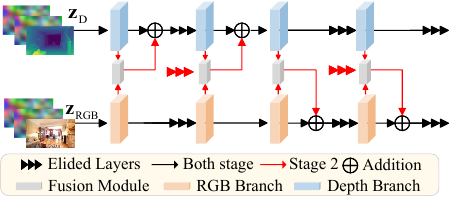}
    \vspace{-1em}
    \caption{
        \textbf{Training scheme of DualCamCtrl.} 
        The overall training follows a two-stage training scheme: (I) In the {decouple stage} (denoted by only the black arrows), the RGB and depth branches are trained separately (with no interaction) to learn modality-specific representations. (II) In the {fusion stage} (both red and black arrows). In this process, SIGMA mechanism is applied through a 3D-aware fusion module, enabling effective cross-modal integration and improved synthesis consistency.
    }
    \vspace{-1em}
    \label{fig:openreels_train_stage}
\end{figure}

\subsection{Detail of Training Pipeline}

As illustrated in~\cref{fig:openreels_train_stage}, DualCamCtrl is trained in two stages. In the decoupled stage, the RGB and depth branches are optimized independently, to acquire modality-specific representations for appearance and geometry. In fusion stage, we enable cross-modal learning by coupling the two branches through a 3D-aware fusion module. This staged strategy lets each branch first stabilize within its own domain and then benefit from complementary cues, leading to stronger cross-modal correspondence.

\subsection{Detail of T2V Setting}

\begin{figure}[htbp]
    \centering
    \includegraphics[width=\linewidth]{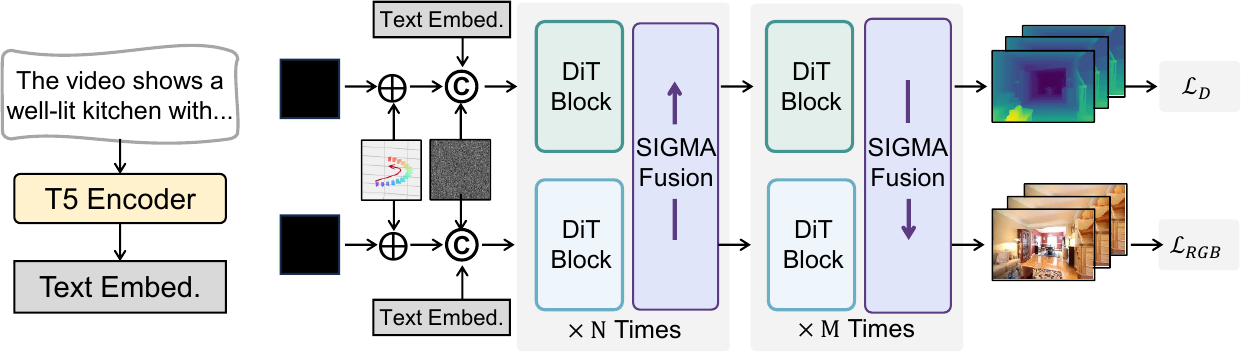}
    \vspace{-1em}
    \caption{
        \textbf{Architecture of DualCamCtrl under the T2V setting.}  The model operates without any visual conditioning. The textual input is first encoded by a T5 encoder and fed into two diffusion transformers to condition both RGB and depth generation.  
    }
    \vspace{-1em}
    \label{fig:t2v_arch}
\end{figure}

Under the T2V setting, the model still operate in a dual-branch jointly generation manner but without any input image or initial frame as conditioning information~(\cref{fig:t2v_arch}). Nevertheless, our strategy allows the RGB branch to effectively provide semantic guidance to the depth branch during generation. We observe that the overall performance consistently improves~(\cref{table:benchmarkcomp_i2v_combined,table:benchmarkcomp_t2v_combined}), indicating the robustness of our cross-modal guidance mechanism.

\section{Findings and Analysis}
\label{appendix:findings_and_analysis}
This section highlights several key findings observed throughout our experiments and provides further analysis to better understand the behavior and characteristics of the proposed framework.

\subsection{Initial Interpretation of Depth Information}

\begin{figure}[htbp]
    \centering
    \includegraphics[width=\linewidth]{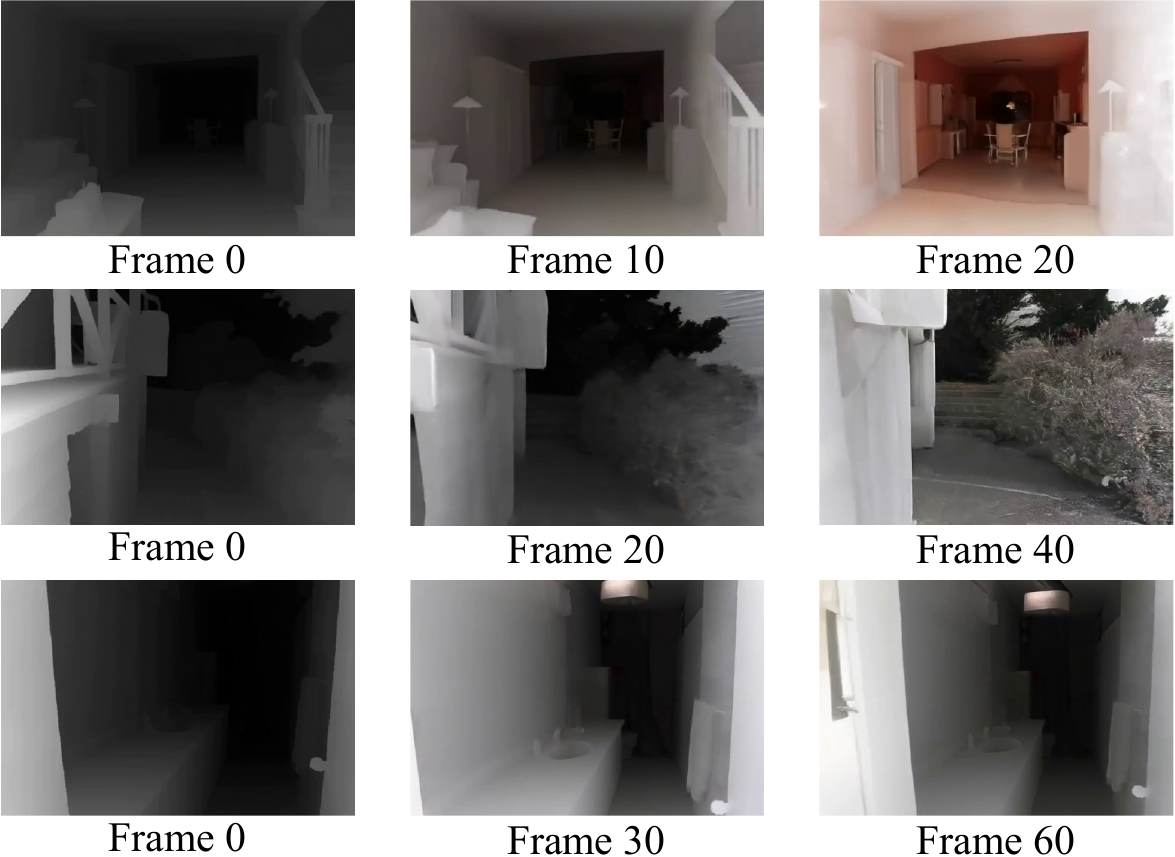}
    \vspace{-1.2em}
    \caption{\textbf{Illustration of \textit{depth-conditioned generation}}. The depth-conditioned generation may initially be processed similarly to low-light or hazy RGB input. The resulting videos maintain structural coherence, reflecting cross-modal adaptability.
    }
    \vspace{-1em}
    \label{fig:misleading_generation}
\end{figure}

Under our dual-branch framework, depth information is initially provided through the RGB branch in the first stage, where the depth map serves as the conditioning input for the first frame. This raises an interesting question: \emph{how is the depth map initially interpreted during this process?}

We observe that directly providing a depth map as the input condition can lead the model to process it similarly to low-light, hazy, or monochromatic visual cues, rather than as a fully distinct modality. Nevertheless, the generated videos show minimal distortion or artifacts, indicating that the current state-of-the-art VAE-DiT~(\cite{wan2025123}) backbone exhibits strong robustness and notable adaptability across input domains~(\cref{fig:misleading_generation}).

\subsection{Extented Analysis}

\begin{figure}[htbp]
    \centering
    \includegraphics[width=\linewidth]{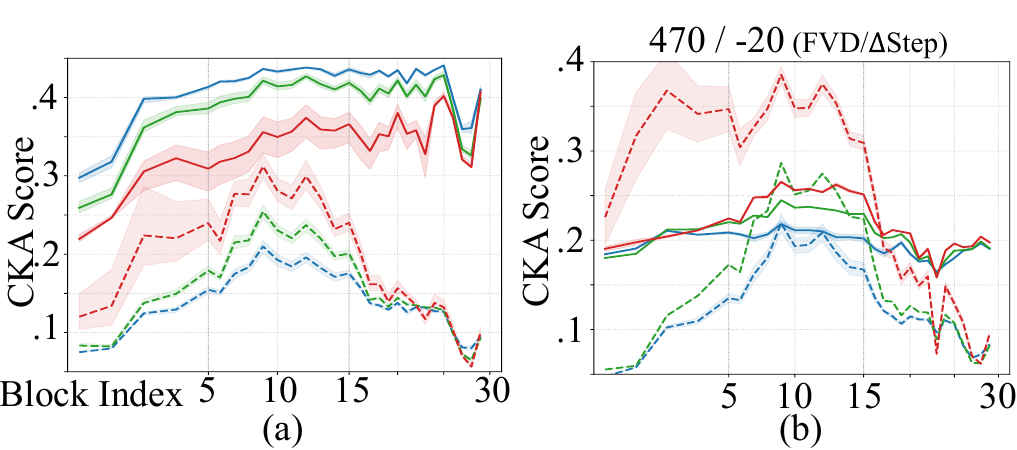}
    \vspace{-1em}
    \caption{\textbf{CKA analysis of depth evolution during denoising; importance of early stage denoising.} (a) Depth branch exhibits a gradual and persistent increase in similarity, indicating continuous geometric refinement and improved spatial consistency as denoising progresses.     (b) Insufficient early-stage denoising results in high variance of CKA similarity across steps, suggesting unstable feature alignment. This instability propagates to later stages, degrading overall spatial coherence and FVD performance.}
    \vspace{-1em}
    \label{fig:depth_analysis_and_fvd_degrade}
\end{figure}

\paragraph{How Depth Latent Evolves?} We further investigate how the depth branch becomes involved and evolves under the guidance of Plücker embeddings~(\cref{fig:depth_analysis_and_fvd_degrade}). It's shown that the depth branch exhibits a steadily increasing similarity that remains high throughout denoising, especially during the {middle and late stages}. This indicates that depth features continuously absorb and reinforce geometric cues to maintain spatial fidelity. Unlike RGB, which relies on early global alignment, depth evolves gradually and persistently contributes to local geometric correction. This also suggests that the conditioning for the depth branch should remain active across the entire process to sustain precise geometric reconstruction.

\vspace{-1em}
\paragraph{Effect of Insufficient Early Stage Denoising.}
As discussed above, increasing the number of early denoising steps effectively improves alignment stability and overall video quality.
We further experimented with the opposite setting and found that insufficient early-stage denoising leads to unstable alignment and degraded FVD performance, as reflected by the high variance in CKA similarity in~\cref{fig:depth_analysis_and_fvd_degrade}.
This suggests that the camera–scene relationship is primarily established at the beginning of the diffusion process, and insufficient early refinement disrupts later-stage feature consistency.

\begin{figure}[htbp]
    \centering
    \includegraphics[width=\linewidth]{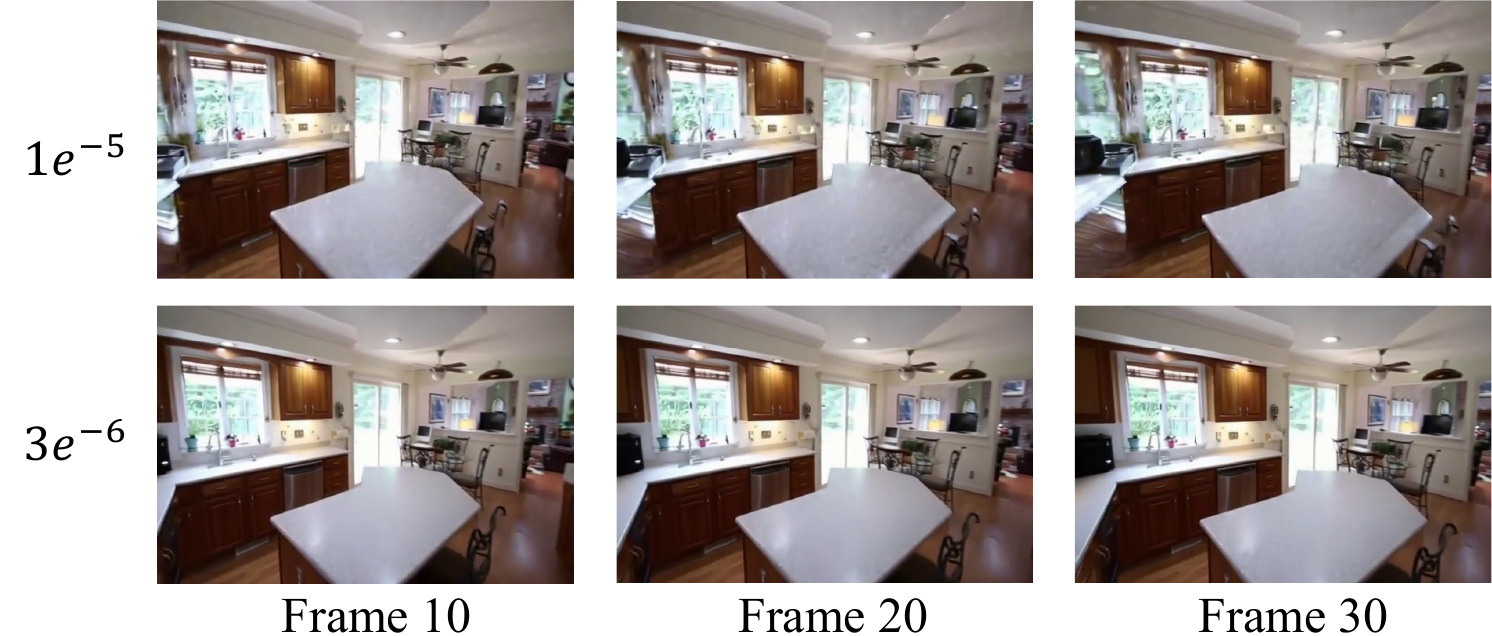}
    \vspace{-1em}
    \caption{\textbf{Comparison of learning rate.} High learning rate~(\ie $1e^{-5}$) lead to unstable training and fail to converge properly. Please zoom in for better visualization.}
    \label{fig:compare_lr}
    \vspace{-0.4em}
\end{figure}

\begin{figure*}[t]
    \centering
    \includegraphics[width=\textwidth]{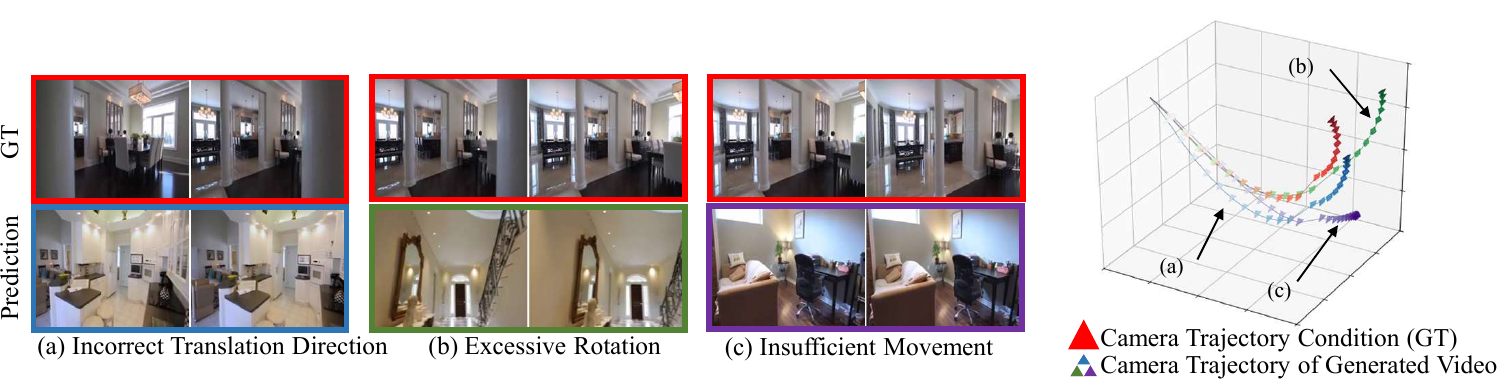} 
    \captionof{figure}{\textbf{Solely Relying on Pl\"ucker Embedding Lacks Scene Understanding.} Given the same camera motion, variations in camera movement are observed in the generated video by~\cite{wan2025123} across different scenes, exhibiting distinct behaviors shown in (a), (b), and (c).}
    \label{fig:limitation}
\end{figure*}

\begin{table*}[htbp]
\centering
\caption{\textbf{User Study Results.} Normalized scores (0--1, higher is better) averaged across participants. Each group was evaluated independently to ensure fairness under different numbers of compared methods.}
\vspace{-1em}
\label{table:user_study_results}
\vspace{0.5em}
\renewcommand{\arraystretch}{0.8}
\begin{tabular}[width=\linewidth]{lccccc}
\toprule
    \textbf{Method} & \textbf{Consistency} & \textbf{Smoothness} & \textbf{Visual Quality} & \textbf{Semantic Consistency} & \textbf{Average} \\
    \midrule
    \multicolumn{6}{l}{\textit{Image-to-Video Setting}} \\
    \midrule
    Ours & \textbf{0.96} & \textbf{0.95} & \textbf{0.94} & \textbf{0.97} & \textbf{0.96} \\
    Wans~\cite{wang2024motionctrl} & 0.88 & 0.87 & 0.86 & 0.88 & 0.87 \\
    SVC~\cite{zhou2025stable} & 0.80 & 0.79 & 0.78 & 0.81 & 0.80 \\
    CameraCtrl~\cite{he2025cameractrl} & 0.77 & 0.76 & 0.75 & 0.78 & 0.77 \\
    MotionCtrl~\cite{wang2024motionctrl} & 0.74 & 0.73 & 0.72 & 0.75 & 0.74 \\
    \midrule
    \multicolumn{6}{l}{\textit{Text-to-Video Setting}} \\
    \midrule
    Ours & \textbf{0.94} & \textbf{0.93} & \textbf{0.92} & \textbf{0.95} & \textbf{0.94} \\
    AC3D~\cite{bahmani2025ac3d} & 0.86 & 0.84 & 0.83 & 0.87 & 0.85 \\
    CameraCtrl~\cite{he2025cameractrl} & 0.79 & 0.78 & 0.77 & 0.80 & 0.78 \\
    MotionCtrl~\cite{wang2024motionctrl} & 0.76 & 0.75 & 0.74 & 0.77 & 0.75 \\
\bottomrule
\end{tabular}
\end{table*}

\section{Additional Implementation Details}
\label{sec:data-preprocessing}

\paragraph{Ray Conditioning.} 
For each camera view, we generate rays by back-projecting pixel coordinates into 3D using the camera intrinsics and extrinsics. 
Specifically, given the camera-to-world transformation $c2w$ and intrinsics $K = (f_x, f_y, c_x, c_y)$, we compute normalized ray directions in camera space and transform them to world coordinates. 
Each ray is represented in the Plücker form $[\mathbf{m}, \mathbf{d}]$, where $\mathbf{d}$ denotes the ray direction and $\mathbf{m} = \mathbf{o} \times \mathbf{d}$ encodes the corresponding moment. 
This process yields a dense field of Plücker coordinates for all pixels across all views~(\cref{alg:ray_condition}). We then encode these Plücker rays using a pre-trained \texttt{WANv2.1} camera pose encoder~\cite{wan2025123}, which produces compact latent features that capture view-dependent geometry. 
Specifically, it applys an 8$\times$ downsampling operation implemented via pixel shuffle~{\cite{shi2016real}} to align with the spatial resolution of the VAE encoder.

\vspace{-1em}
\paragraph{Hyperparameter.} We used a batch size of 8. Compared to the original training setup~\cite{wan2025123}, which adopted a learning rate of $1\times10^{-5}$, we used a smaller learning rate of $3\times10^{-6}$. The learning rate scheduler remained linear without any modification. We found that this configuration led to better convergence performance~(\cref{fig:compare_lr}).

\begin{figure*}[t]
    \centering
    \includegraphics[width=\linewidth]{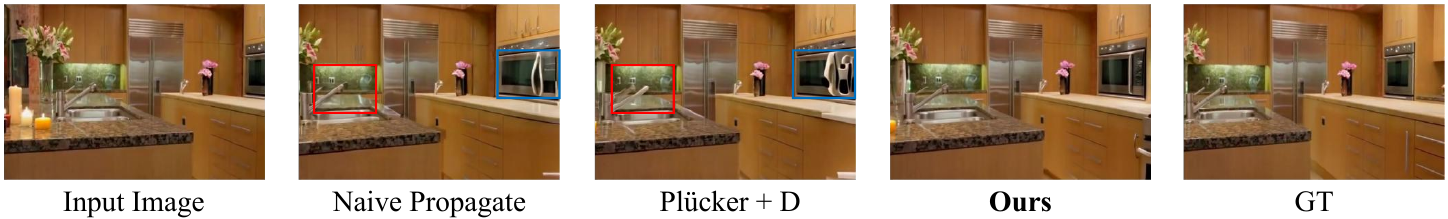}
    \vspace{-1em}
    \caption{\textbf{Comparison of fusion strategy.} Single-frame depth conditioning fails to maintain a coherent understanding across the entire video, resulting in artifacts and unnatural generation. In contrast, our multi-frame depth synthesis methods help produce more plausible outputs. Please zoom in for more details.}
\label{fig:single_frame_depth}
    \vspace{-1em}
\end{figure*}

\vspace{-1em}
\paragraph{Evaluation Datasets.} We evaluate our methods on two datasets: \textsc{RE10K}~\cite{46965} and \textsc{DL3DV}~\cite{ling2024dl3dv}. For \textsc{RE10K}, we randomly sample frames with a stride of 2–8. For \textsc{DL3DV}, which already exhibits large camera motion, we use a stride of 1. We use VBench~\cite{huang2024vbench} to exclude videos whose motion-smoothness and consistency score is low.

\vspace{-1em}
\paragraph{Prompt Generation.}~We generate prompts for every sampled video to provide richer supervisory signals. However, generating captions for all frames can be computationally expensive and resource-inefficient, due to substantial frame overlap within each video. To mitigate this, we generate captions for frames every chosen sample stride, ensuring that each video contains at least one frame with an associated caption while avoiding redundant computation.

\section{Additional Experiments}

\subsection{More Experimental Analysis}

\paragraph{Limitation of sole reliance on pl\"ucker embedding.}
We found that although recent camera-conditioned generative models can recover or specify camera trajectories, they still exhibit important limitations. Many approaches encode motion in Plücker  embeddings without explicit reasoning about scene geometry or semantics, causing the model to conflate camera motion with content-dependent appearance. As a result, the same nominal trajectory yields scene-specific behaviors, like changes in unintended rotations, and inconsistent parallax as illustrated in Fig.~\ref{fig:limitation}.

\vspace{-1em}
\paragraph{User Study.}~In our user study, we asked participants to evaluate the following aspects of the generated videos~(\cref{table:user_study_results}):
\textbf{Consistency of Camera Trajectories with GT:} How closely do the camera trajectories match the expected ground truth, maintaining the correct motion paths?    
\textbf{Smoothness and Coherence:} How smooth and continuous is the video, without any noticeable jerks, flickering, or temporal discontinuities?
\textbf{Visual Quality and Detail:} How high is the overall visual quality and detail of the rendered frames?
 \textbf{Semantic Consistency with GT:} How well does the generated scene match the expected semantic content in terms of object recognition, spatial relationships, and scene structure? The user study results (Table~\cref{table:user_study_results}) show that our method outperforms all baselines across all four evaluation aspects.

\begin{figure}[htbp]
    \centering
    \includegraphics[width=\linewidth]{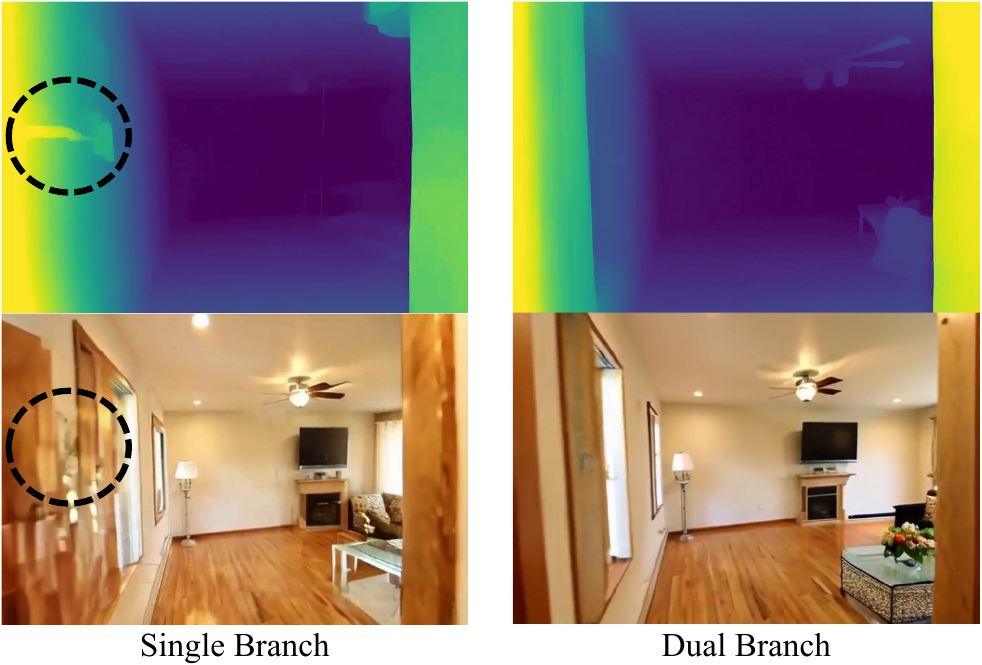}
    \vspace{-1em}
    \caption{\textbf{Illustration of coupling the RGB and depth modalities within a single branch leads to interference.} This interference causes misalignment between the two modalities and adversely affecting the generation quality.}
    \label{fig:single_branch_misalign}
\end{figure}

\vspace{-1em}
\paragraph{Comparison of Inject Depth  Strategies}
To demonstrate that achieving coherent interaction between depth and RGB under explicit camera control is non-trivial, we compare several integration strategies~(\cref{fig:single_frame_depth}):

\begin{algorithm}[htbp]
    \caption{Ray Condition: Plücker Ray Generation}
    \label{alg:ray_condition}
    \begin{algorithmic}[1]
    \Require Camera intrinsics $K \in \mathbb{R}^{B \times V \times 4}$, 
    camera-to-world matrices $c2w \in \mathbb{R}^{B \times V \times 4 \times 4}$, 
    image size $(H, W)$, 
    \Ensure Plücker ray representation $\text{plücker} \in \mathbb{R}^{B \times V \times H \times W \times 6}$
    
    \State Initialize pixel grids $(i, j) \leftarrow \text{meshgrid}(0, \ldots, W-1; 0, \ldots, H-1)$
    \For{each batch $b = 1, \ldots, B$}
        \For{each view $v = 1, \ldots, V$}
   
            \State Extract intrinsics $(f_x, f_y, c_x, c_y) \leftarrow K[b, v]$
    
            \State Compute camera-space directions:
            \[
            x = \frac{i - c_x}{f_x}, \quad 
            y = \frac{j - c_y}{f_y}, \quad 
            z = 1
            \]
            \State Normalize direction vectors: 
            $\mathbf{d}_{cam} = \text{normalize}([x, y, z])$
    
            \State Transform to world coordinates:
            \[
            \mathbf{d}_{world} = R_{b,v} \mathbf{d}_{cam}, \quad
            \mathbf{o}_{world} = t_{b,v}
            \]
    
            \State Expand $\mathbf{o}_{world}$ to match the resolution $(H \times W)$
    
            \State Compute Plücker coordinates:
            \[
            \mathbf{m} = \mathbf{o}_{world} \times \mathbf{d}_{world}, \quad
            \text{plücker}[b,v] = [\mathbf{m}, \mathbf{d}_{world}]
            \]
        \EndFor
    \EndFor
    \State \Return $\text{plücker}$
    \end{algorithmic}
    \end{algorithm}

\begin{figure*}[t]
    \centering
    \includegraphics[width=\linewidth]{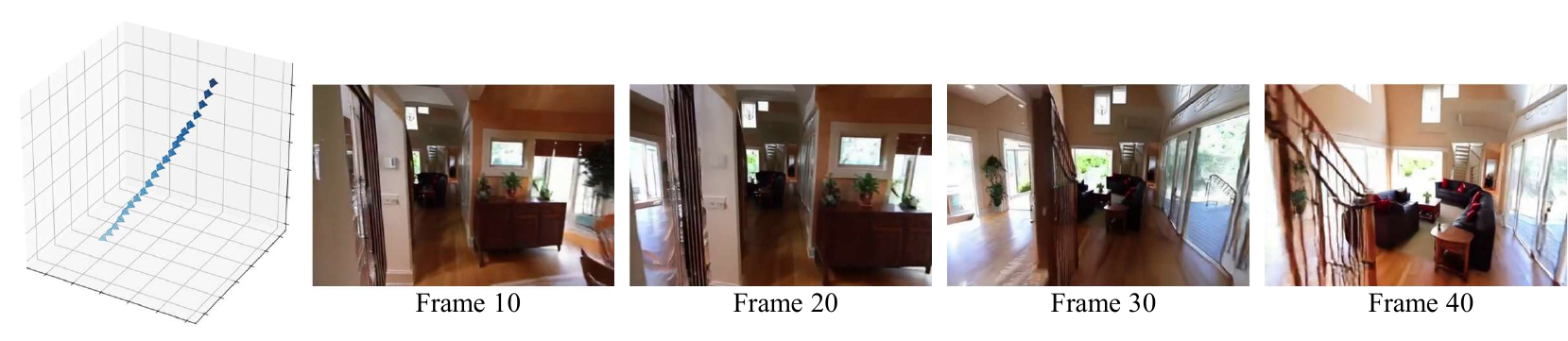}
    \vspace{-1.5em}
    \caption{\textbf{Failure case under large motion.} When the inter-frame motion is large, severe artifacts emerge. Please zoom in for details.}
    \label{fig:large_motion_failure}
    \vspace{-1em}
\end{figure*}

\vspace{-1em}
\paragraph{Single-frame depth conditioning} treats depth as static or simply propagates it across frames. In detail, we compare two strategies: 
\begin{enumerate}
    \item \textbf{Naive-Prop}: propagate the depth of the first frame to all subsequent frames and use it as a conditioning signal~(\ie use a side branch but with static frame).
    \item \textbf{Pl\"ucker + D}: incorporate depth as an additional dimension in the Pl\"ucker ray embedding.
\end{enumerate}

\vspace{-1em}
\paragraph{Single-branch integration} adds an auxiliary depth branch with a shared encoder, generating RGB and depth within a single branch, similar with~\cite{wang2025transpixeler}.

As illustrated in~(\cref{fig:single_frame_depth,fig:single_branch_misalign,table:ablation_integration}), single-frame depth conditioning imposes a static prior and frequently introduces visible artifacts across time, leading to unnatural generations. In contrast, single-branch integration couples RGB and depth within a single pathway and causes unacceptable cross-modal interference, further exacerbating misalignment. By comparison, our dual-branch framework enables balanced interaction without entanglement, achieving consistent geometry and vivid appearance.


\subsection{More ablation study}
\label{appendix:more_ablation}
\paragraph{Effect of 3D fusion strategy.}
We perform a progressive ablation to evaluate each component of the 3D fusion block.
As shown in \cref{table:ablation_dsf}, adding spatial, 3D spatiotemporal, and frame-gating modules progressively improves performance.

\begin{table}[t]
    \centering\small
    \caption{\textbf{Ablation on integration strategy.}}
    \setlength{\tabcolsep}{4pt}    
    \resizebox{1\linewidth}{!}{
    \begin{tabular}{l|cc|cc}
        \toprule
        Integration Strategy & FVD $\downarrow$ & FID $\downarrow$ & MS $\uparrow$ & FC $\uparrow$ \\
        \midrule
            Naive-Prop   & 98.6 & 62.7 & 0.9823 & 0.9589 \\
        Pl\"ucker + D  & 100.8 & 65.2 & 0.9891 & 0.9648 \\
        \midrule
        Single-Branch & 125.1 & 66.0 & 0.9787 & 0.9515 \\
        \midrule

        Ours  & \hllfirst{80.4} & \hllfirst{49.9} & \hllfirst{0.9886} & \hllfirst{0.9677} \\
        \bottomrule
    \end{tabular}
    }
    \label{table:ablation_integration}
\end{table}

\begin{table}[t]
    \centering\small
    \caption{\textbf{Ablation on 3D fusion block.}}
    \vspace{-1em}
    \setlength{\tabcolsep}{4pt}    
    \resizebox{1\linewidth}{!}{
    \begin{tabular}{l|cc|cc}
        \toprule
        Variant & FVD $\downarrow$ & FID $\downarrow$ & MS $\uparrow$ & FC $\uparrow$ \\
        \midrule
        ControlNet-Style (1D)  & 112.3 & 61.8 & 0.9819 & 0.9641 \\
        Spatial (2D) Fusion  & {98.4} & {60.1} & {0.9815} & {0.9643} \\
        Spatial Temporal (3D) Fusion & \hllsecond{85.4} & \hllsecond{52.7} & \hllsecond{0.9861} & \hllsecond{0.9648} \\
        3D Fusion + Frame Gating  & \hllfirst{80.4} & \hllfirst{49.9} & \hllfirst{0.9886} & \hllfirst{0.9677} \\
        \bottomrule
    \end{tabular}
    }
    \label{table:ablation_dsf}
\end{table}

\vspace{-1em}
\paragraph{Effect of {{S}}emat{{i}}c {{G}}uided {{M}}utual {{A}}lignment.}
We further investigate the impact of cross-branch communication and fusion depth. Three fusion strategies are compared:
(1) \textit{One-way Alignment},
(2) \textit{Geometry-Guided Alignment},
(3) \textit{{{S}}emat{{i}}c {{G}}uided {{M}}utual {{A}}lignment. (Ours)},
In \cref{tab:ablation_pgra}, the left two columns indicate the layer ranges where RGB and depth features are injected, allowing us to analyze the effect of fusion depth under each strategy.
As shown in \cref{tab:ablation_pgra}, SIGMA consistently achieves the best video quanlity metrics as well as camera motion consistency, indicating that semantic guided mutual alignment is more effective than one-way or geometry-guided conditioning.
\begin{table}[t]
    \centering\small
    \caption{\textbf{Analysis on different alignment mechanism.}}
    \vspace{-0.7em}
    \setlength{\tabcolsep}{4pt}
        \resizebox{1\linewidth}{!}{
    \begin{tabular}{cc|cc|cc|cc}
        \toprule
        RGB & Depth & FVD $\downarrow$ & FID $\downarrow$ & MS $\uparrow$ & FC $\uparrow$ & \sre $\downarrow$ & \ste $\downarrow$ \\
        \midrule
        - & 1-5   & 91.4 & 59.1 & 0.984 & {0.966} & 1.76 & 0.36 \\
        - & 1-10  & 87.0 & 58.4 & 0.984 & 0.964         & 1.56 & 0.37 \\
        - & 1-15  & 83.2 & 59.7 & 0.985 & 0.964         & {1.55} & 0.35 \\
        \midrule
        6-10 & 1-5   & 87.4 & 54.2 & 0.985 & {0.966} & 1.56 & {0.33} \\
        6-15 & 1-5   & 84.6 & 53.2 & \hllsecond{0.987} & {0.966} & {1.55} & {0.34} \\
        11-15& 1-10  & 84.2 & 53.7 & \hllsecond{0.987} & 0.965 & {1.55} & {0.34} \\
        \midrule
        1-5  & 6-10  & {80.9} & \hllfirst{49.5} & \hllsecond{0.987} & 0.965 & \hllsecond{1.27} & \hllsecond{0.25} \\
        1-5  & 6-15  & \hllfirst{80.4} & \hllsecond{49.9} & \hllfirst{0.989} & \hllsecond{0.968} & \hllfirst{1.25} & \hllsecond{0.25} \\
        1-10 & 11-15 & \hllsecond{80.3} & {50.2} & \hllfirst{0.989} & \hllfirst{0.969} & \hllsecond{1.27} & \hllfirst{0.23} \\
        \bottomrule
    \end{tabular} }
    \label{tab:ablation_pgra}
\end{table}


\subsection{More Visual Comparison}
\begin{figure*}[!b]
    \centering
    \includegraphics[width=0.96\linewidth]{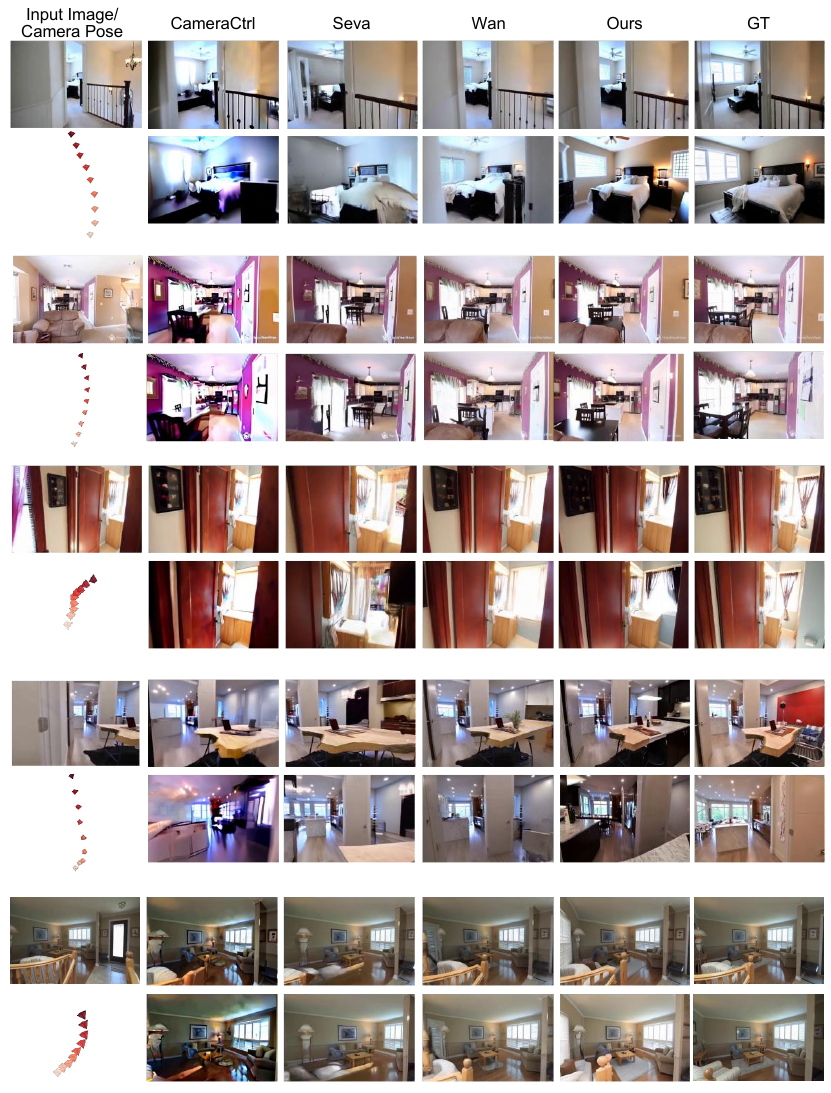}
    \vspace{-15pt} 
    \caption{\textbf{More qualitative results of DualCamCtrl.} Please zoom in for better visualization.}
    \label{fig:visualization_2}
\end{figure*}

\begin{figure*}[!b]
    \centering
    \includegraphics[width=0.96\linewidth]{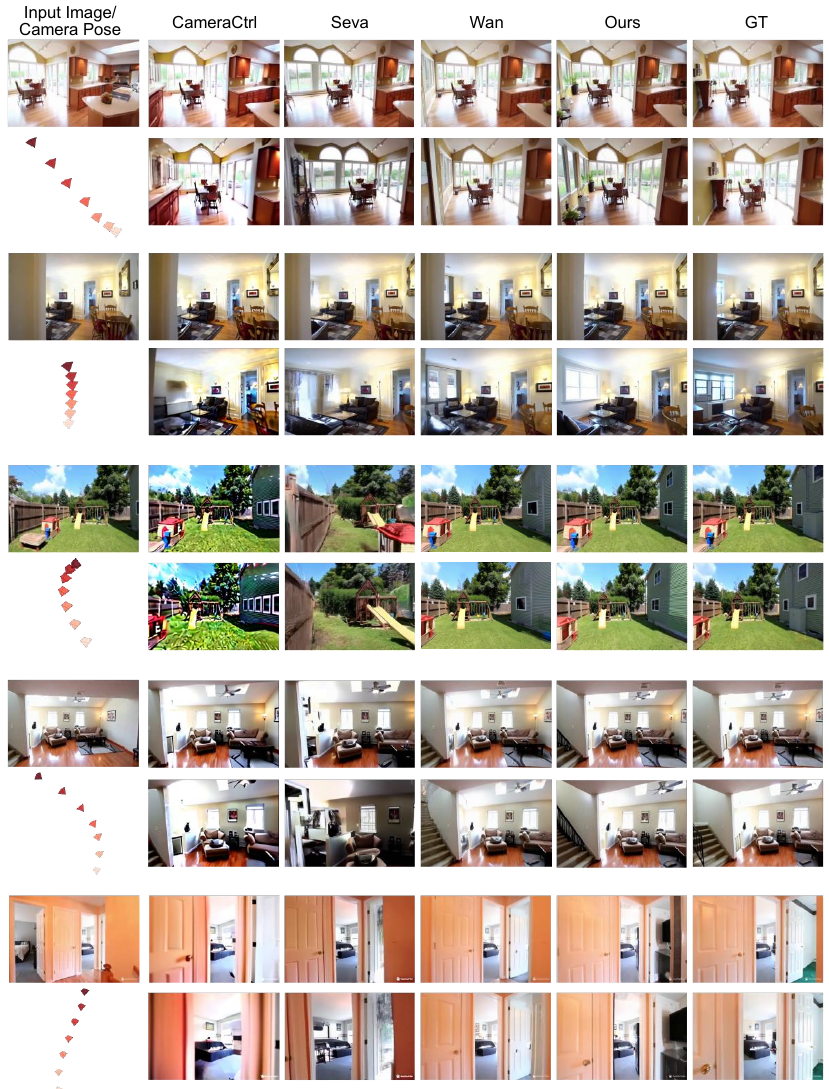}
    \vspace{-15pt} 
    \caption{\textbf{More qualitative results of DualCamCtrl.} Please zoom in for better visualization.}
    \label{fig:visualization_1}
\end{figure*}

We provide additional examples to further demonstrate the effectiveness of our method~(\cref{fig:visualization_1,fig:visualization_2}).


\section{Discussion and Limitation}
\label{sec:limitation}

While our method performs strongly, we identify two intriguing aspects that present potential avenues for enhancement. We discuss them below with the aim of stimulating future research.

\vspace{-1em}
\paragraph{Large Motion.}
Our method also suffers from issues related to large motion, which is a common challenge in video generation tasks. Other methods use NVS to resolve ambiguities and mitigating flickering for video  generation~\cite{yu2024viewcrafter,ren2025gen3c,liu2024reconx,gao2024cat3d,wu2024reconfusion}. However, despite our advancements, large motion still presents a difficulty in stable and coherent end to end video generation~(\cref{fig:large_motion_failure}).
\vspace{-1em}
\paragraph{Model Capacity and Parameter Efficiency.}
Finally, although our parameter count remains comparable to prior designs~\cite{gu2025das,feng2024i2vcontrol,he2024cameractrl}, the overhead in our framework mainly arises from duplicating the backbone to form separate RGB and depth branches. This duplication is acceptable at the 1.3B scale, where even two copies of the \textsc{Wan V2.1} backbone still yield clear performance advantages, but becomes less favorable when scaling to larger video diffusion models. A promising future direction is to develop more parameter efficient depth guidance, for instance by distilling a compact depth branch that can effectively guide the RGB pathway while keeping the overall model lightweight.
\clearpage

\end{document}